\documentclass{article} 
\usepackage{arxiv}
\usepackage{natbib}

\date{}
\renewcommand{\headeright}{}
\renewcommand{\undertitle}{}
\renewcommand{\shorttitle}{\textit{}}

\usepackage{amsmath,amsfonts,bm}

\def\eqref#1{equation~\ref{#1}}

\def\1{\bm{1}}

\DeclareMathAlphabet{\mathsfit}{\encodingdefault}{\sfdefault}{m}{sl}
\SetMathAlphabet{\mathsfit}{bold}{\encodingdefault}{\sfdefault}{bx}{n}

\usepackage{graphicx}
\usepackage{hyperref}
\usepackage{url}
\usepackage{comment}
\usepackage{multicol}
\usepackage{multirow}
\usepackage{booktabs}
\usepackage{wrapfig}

\usepackage{pgfplots}
\usetikzlibrary{pgfplots.groupplots}
\pgfplotsset{compat=1.3}
\usepackage{tikz}
\usetikzlibrary{patterns}
\usepackage{pgf-pie}

\definecolor{battleshipgrey}{rgb}{0.3, 0.3, 0.3}
\definecolor{brilliantrose}{rgb}{1.0, 0.33, 0.64}
\definecolor{americanrose}{rgb}{1.0, 0.01, 0.24}
\definecolor{jweigreen}{rgb}{0,0.45,0.24}
\definecolor{bluegray}{rgb}{0.1, 0.1, 0.4}
\definecolor{ao(english)}{rgb}{0.0, 0.5, 0.0}
\definecolor{blanchedalmond}{rgb}{1.0, 0.92, 0.8}
\definecolor{atomictangerine}{rgb}{1.0, 0.6, 0.4}
\definecolor{chocolate(web)}{rgb}{0.82, 0.41, 0.12}
\definecolor{bananayellow}{rgb}{1.0, 0.88, 0.21}
\definecolor{goldenbrown}{rgb}{0.6, 0.4, 0.08}
\definecolor{aliceblue}{rgb}{0.94, 0.97, 1.0}
\definecolor{beige}{rgb}{0.96, 0.96, 0.86}
\definecolor{babyblue}{rgb}{0.54, 0.81, 0.94}
\definecolor{camel}{rgb}{0.76, 0.6, 0.42}
\definecolor{cinnamon}{rgb}{0.82, 0.41, 0.12}

\usepackage{pgfplots}
\usetikzlibrary{pgfplots.groupplots}
\pgfplotsset{compat=1.3}
\usepackage{pifont}

\title{Parameter-Efficient Fine-Tuning Design Spaces}

\author{Jiaao Chen$^\dagger$\thanks{Work done during an internship at Amazon Web Services. Correspondence to Jiaao Chen $\textlangle$jiaaochen@gatech.edu$\textrangle$ and Aston Zhang $\textlangle$astonz@amazon.com$\textrangle$.}, Aston Zhang$^\ddagger$, Xingjian Shi$^\ddagger$, Mu Li$^\ddagger$, Alex Smola$^\ddagger$, Diyi Yang$^\diamond$\\
$^\dagger$Georgia Institute of Technology, $^\ddagger$Amazon Web Services, $^\diamond$Stanford University
}

\begin{document}

\maketitle

\setcounter{footnote}{0}

\begin{abstract}

Parameter-efficient fine-tuning aims to achieve performance comparable to fine-tuning, using fewer trainable parameters. Several strategies (e.g., Adapters, prefix tuning, BitFit, and LoRA) have been proposed. However, their designs are hand-crafted separately, and it remains unclear whether certain design patterns exist for parameter-efficient fine-tuning. Thus, we present a parameter-efficient fine-tuning design paradigm and discover design patterns that are applicable to different experimental settings. Instead of focusing on designing another individual tuning strategy, we introduce parameter-efficient fine-tuning design spaces that parameterize  tuning structures and tuning strategies. Specifically, any design space is characterized by four components: layer grouping, trainable parameter allocation, tunable groups, and strategy assignment.
Starting from an initial design space,
we progressively refine the space based on the model quality of each design choice 
and make greedy selection at each stage over these four components.
We discover the following design patterns:
(i) group layers in a spindle pattern; 
(ii) allocate the number of trainable parameters to layers uniformly;
(iii) tune all the groups; 
(iv) assign proper tuning strategies to different groups.
These design patterns result in new parameter-efficient fine-tuning methods. We show experimentally that these methods consistently and significantly outperform investigated parameter-efficient fine-tuning strategies across different backbone models and different tasks in natural language processing\footnote{Code is available at: \url{https://github.com/amazon-science/peft-design-spaces}.}. 

\end{abstract}

\section{Introduction}
Large pretrained models have achieved the state-of-the-art performances across a wide variety of downstream natural language processing tasks through fine-tuning on task-specific labeled data \citep{devlin2018bert, liu2019roberta, yang2019xlnet, joshi2019spanbert, sun2019ernie, clark2019electra, lewis2019bart, bao2020unilmv2, he2020deberta, raffel2020exploring, ziems-etal-2022-value}. However, fine-tuning all the parameters and storing them separately for different tasks is expensive in terms of computation and storage overhead (e.g., $355$M parameters for RoBERTa \citep{liu2019roberta} and $175$B parameters for GPT-$3$ \citep{brown2020language}). This makes it difficult to deploy in real-world natural language processing (NLP) systems composed of multiple tasks.

To adapt general knowledge in pretrained models to specific down-stream tasks in a more parameter-efficient way, various strategies have been proposed where only a small number of (extra) parameters are learned while the remaining pretrained parameters are frozen \citep{pmlr-v97-houlsby19a,pfeiffer-etal-2021-adapterfusion,li2021Prefixtuning,brown2020language,lester2021power,schick-schutze-2021-exploiting,ziems-etal-2022-value}.
Adapter tuning \citep{pmlr-v97-houlsby19a} is among the earliest strategies to steer pretrained models with a limited number of parameters. It inserts adapters (small neural modules) to each layer of the pretrained network and only the adapters are trained at the fine-tuning time. 
Inspired by the success of prompting methods that control pretrained language models through textual prompts \citep{brown2020language}, prefix tuning \citep{li2021Prefixtuning} and prompt tuning \citep{lester-etal-2021-power} prepend additional tunable tokens to the input or hidden layers and only train these soft prompts when fine-tuning on downstream tasks. BitFit \citep{https://doi.org/10.48550/arxiv.2106.10199} updates the bias terms in pretrained models while freezing the remaining parameters. LoRA \citep{https://doi.org/10.48550/arxiv.2106.09685} decomposes attention weight gradients into low-rank matrices to reduce the number of trainable parameters. 
With promising results from such research,
\citet{https://doi.org/10.48550/arxiv.2110.04366} proposed a unified view of these existing strategies and illustrated differences and connections among them. Like its antecedents, the resulting method is still \emph{equally} assigned to different pretrained layers.

Despite being effective, most parameter-efficient fine-tuning strategies have been developed via manual design processes, without much consideration of whether design patterns
exist across these different strategies and how such patterns might apply to different backbone models and downstream tasks. Moreover, different strategies are usually applied separately; thus, it is unclear which strategy works best when and where \citep{mao-etal-2022-unipelt}, as well as how these different strategies reinforce or complement each other. In this light, our goal is to understand the parameter-efficient fine-tuning design in a more comprehensive view and discover design patterns that are both interpretable and applicable across different experimental settings. 

\begin{figure}[t]
\centering
\includegraphics[width=0.9\textwidth]{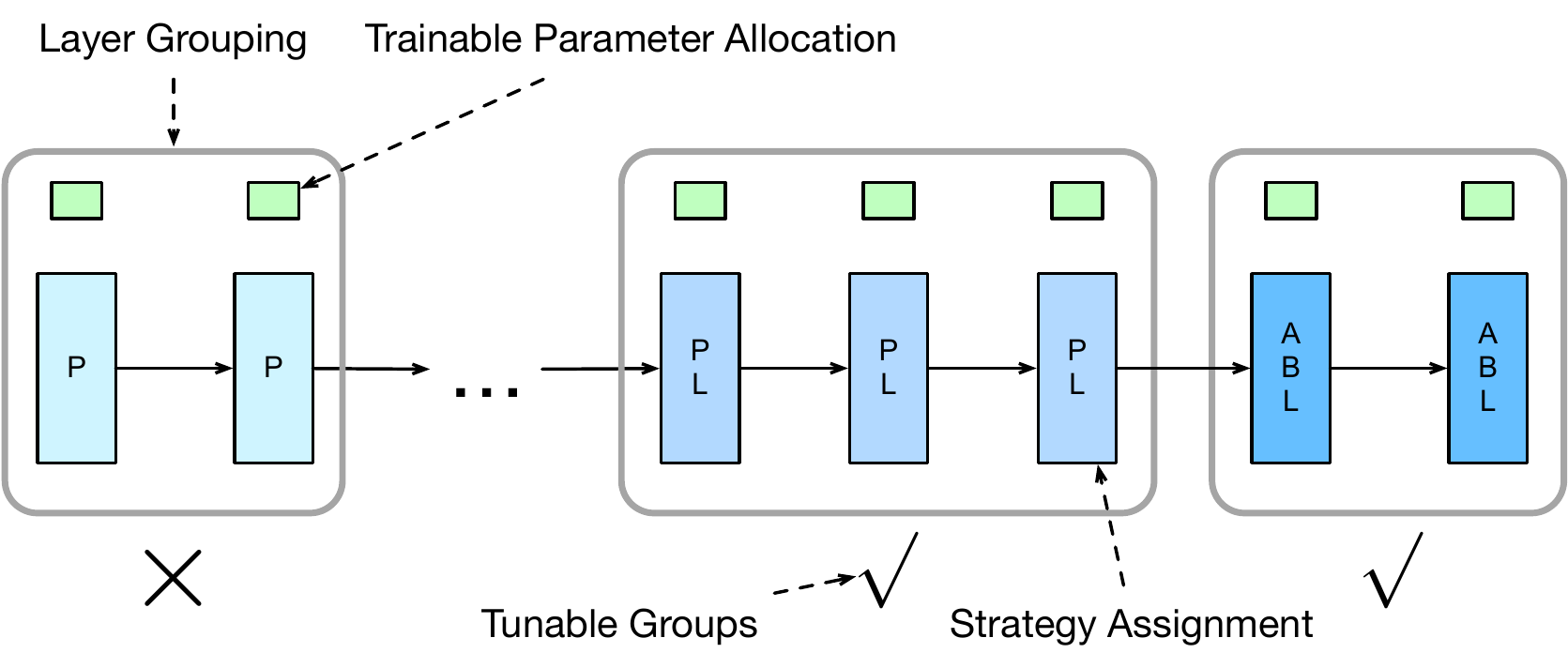}
\caption{A parameter-efficient fine-tuning design space. It is characterized by 
(i) layer grouping (how to group consecutive layers),
(ii) trainable parameter allocation (how to allocate the  number of trainable parameters to layers),
(iii) tunable groups (which groups will be finetuned), 
and (iv) strategy assignment (how to assign proper strategies, such as among \textbf{A}dapter, \textbf{P}refix, \textbf{B}itFit, and \textbf{L}oRA, to groups).}
\label{Fig:design_space}
\end{figure}

Instead of designing yet another individual strategy that is equally applied to different pretrained layers, we introduce \textbf{parameter-efficient fine-tuning design spaces} that parameterize both tuning structures and strategies.
More concretely, any of these design spaces is characterized by four major components as shown in Figure~\ref{Fig:design_space}: \emph{layer grouping}, \emph{trainable parameter allocation}, \emph{tunable groups}, and \emph{strategy assignment}. 


Starting from a relatively unconstrained parameter-efficient fine-tuning design space,
we progressively refine the space by
comparing the overall quality of models randomly sampled
from design spaces enforced with
different constraints (e.g., each group has the same number of layers).
Throughout the experimental process, we discover several design patterns for parameter-efficient fine-tuning, such as 
group layers in a spindle pattern, 
allocate the number of trainable parameters to layers uniformly,
tune all the groups, 
and assign proper tuning strategies to different groups. We further introduce new parameter-efficient fine-tuning methods that adopt all these discovered design patterns. Extensive experiments show that our methods consistently outperform investigated parameter-efficient fine-tuning strategies. Although we use T5 \citep{raffel2020exploring} and classification tasks as the working example, 
we find that our methods with all these discovered design patters are applicable to other backbones (e.g., RoBERTa \citep{liu2019roberta}, BART \citep{lewis-etal-2020-bart}, and XLNet \citep{yang2019xlnet}) and different natural language processing tasks (e.g., summarization, machine translation, and eight SuperGLUE datasets).

Our contributions can be summarized as follows: (i) We introduce parameter-efficient fine-tuning design spaces. (ii) Based on these design spaces, we discover several design patterns in parameter-efficient fine-tuning via comprehensive experiments. (iii) Our discovered design patterns lead to parameter-efficient fine-tuning methods, consistently outperforming investigated parameter-efficient fine-tuning strategies across different backbone
models and different NLP tasks. 

\section{Related Work}
Our work is closely related to and built upon the research about the network design spaces and parameter-efficient fine-tuning. We discuss the connections and differences below.

\paragraph{Network Design Spaces}
A lot of works designed neural network models
via an ad-hoc discovery of new design choices that improve performances \citep{https://doi.org/10.48550/arxiv.1905.13214}, 
such as the use of deeper architectures or residuals.
Recently, there have been works \citep{https://doi.org/10.48550/arxiv.2003.13678,https://doi.org/10.48550/arxiv.2011.08843,https://doi.org/10.48550/arxiv.1905.13214} performing at the design space level to discover new design principles for convolutional neural networks \citep{https://doi.org/10.48550/arxiv.2003.13678} and graph neural networks \citep{https://doi.org/10.48550/arxiv.2011.08843}. 
Inspired by this line of research,
we focus on the design space perspective
to rethink parameter-efficient fine-tuning,
with the goal of discovering design patterns
that are applicable to different experimental settings.

\paragraph{Parameter-Efficient Fine-Tuning for NLP}
As pretrained models grow in size, storing fine-tuned models becomes exceedingly expensive, and fine-tuning becomes infeasible for those without extremely high compute resources. A growing body of research has been devoted to finding parameter-efficient alternatives for adapting large-scale pretrained models with reduced memory and storage costs. \citet{houlsby2019parameter} proposed to adapt large models using bottleneck layers (with skip-connections) between each layer. This idea has been extended in many domains ~\citep{stickland2019bert,pfeiffer2020adapterfusion,rebuffi2017learning,lin2020exploring}. Other works have aimed to avoid introducing additional parameters by identifying and training only a subset of all model parameters \citep{zhao2020masking,guo2020parameter,mallya2018piggyback,radiya2020fine, FISHmask, https://doi.org/10.48550/arxiv.2106.10199}. Recent works also explored the idea of rank decomposition based on parameterized hypercomplex multiplications via the Kronecker product \citep{zhang2021beyond} and 
injecting trainable rank decomposition matrices into each layer \citep{https://doi.org/10.48550/arxiv.2106.09685,karimi2021compacter}. \citet{li2021Prefixtuning} introduced prefix-tuning that prepends a set of prefixes to autoregressive language models or prepends prefixes for both encoders and decoders. The prefix parameters are updated while the pretrained parameters are fixed.  \citet{lester2021power} proposed a similar method, but only added virtual tokens at the embedding layer of large-scale models rather than discrete prompts \citep{https://doi.org/10.48550/arxiv.2205.12548,https://doi.org/10.48550/arxiv.2208.03229}. 
\citet{bari2022spt} proposed semi-parametric prompt tuning that converges more easily, where memory prompts are input-adaptive without the need for tuning.
Recently, \citet{https://doi.org/10.48550/arxiv.2110.04366} and \citet{https://doi.org/10.48550/arxiv.2203.06904} proposed a unified view of the existing parameter-efficient fine-tuning strategies and illustrated the difference and connections among them. \citet{mao-etal-2022-unipelt} also introduced a unified framework to combine different methods through mixture-of-experts.

In contrast to 
these aforementioned works that assign their individual method equally to different pretrained layers, we focus on more general design spaces
of parameter-efficient fine-tuning.
This could provide a more comprehensive view of parameter-efficient fine-tuning in terms of both the tuning structures and tuning strategies. 
Through experiments where we progressively 
refine design spaces,
we discover design patterns for parameter-efficient fine-tuning.

\section{Components of Design Spaces}
\label{sec:Components of Design Spaces}
When defining design spaces of parameter-efficient fine-tuning,
we aim to
cover key design components
and 
provide a representative set of choices in each design component.
Note that 
our goal is not to enumerate all possible design spaces, but to demonstrate how the use of design spaces can help inform parameter-efficient fine-tuning research.

Concretely, in our work, 
the parameter-efficient fine-tuning design spaces 
are formed by a representative set of choices in parameter-efficient fine-tuning, which consists of the following four components:
(i) layer grouping, (ii) trainable parameter allocation,
(iii) tunable groups, and (iv) strategy assignment.
Following the illustrated design space example
in Figure~\ref{Fig:design_space},
we describe these four design components in detail below and will explore their design choices in Section~\ref{Sec:Searching}.

\paragraph{Layer Grouping} Different layers in pretrained models capture different information and behave differently. For example, \citet{jawahar-etal-2019-bert} found that the $\{3,4,5,6,7,9,12\}$-th layers have the most representation power in BERT and every layer captures a different type of information ranging from the surface, syntactic, to the semantic level representation of text. For instance, the 9th layer has predictive power for semantic tasks such as checking random swapping of coordinated clausal conjuncts, while the 3rd layer performs best in surface tasks like predicting sentence length.
Therefore when adapting these pretrained models to downstream tasks,
how to group layers with similar behaviors together is critical to the design and application of proper 
parameter-efficient fine-tuning strategies.
For this design component, we study the patterns of how to group consecutive layers in pretrained models (e.g., transformer layers in T5) during the fine-tuning process.

\paragraph{Trainable Parameter Allocation} 
In parameter-efficient fine-tuning,
the total number of trainable parameters is usually preset, 
such as a small portion of the total number of parameters
in the pretrained models.
We will study different design choices for how to allocate
a predefined number of trainable parameters
to layers.

\paragraph{Tunable Groups} \citet{https://doi.org/10.48550/arxiv.2106.10199} found that not all the parameters need to be tuned during fine-tuning on the downstream tasks. For instance, BitFit \citep{https://doi.org/10.48550/arxiv.2106.10199} only updates the bias parameters in pretrained models while freezing the remaining parameters. Thus, we study which groups need to be learned during parameter-efficient fine-tuning to attain better performances.

\paragraph{Strategy Assignment} In order to improve the parameter efficiency, different sets of strategies \citep{li2021Prefixtuning,lester2021power,pmlr-v97-houlsby19a,https://doi.org/10.48550/arxiv.2106.09685} have been proposed where only a small number of (extra) parameters are tuned and the remaining parameters in these pretrained models are frozen to adapt their general knowledge to specific down-stream tasks.
Inspired by effectiveness of offering architectural flexibility \citep{zhang2021beyond,zhang2021self},
we hypothesize that different groups might benefit from different proper strategies (or combinations) for capturing different types of information.    
More formally,
given a set of individual strategies $\mathcal{A}$ for assignment,
for any group $G_i$,
assign a subset $\mathcal{U}_i \subset \mathcal{A}$ to each layer in $G_i$.

\section{Discovering Design Patterns}\label{Sec:Searching}
Building on these four different design components 
of PEFT design spaces, 
we will start from a relatively unconstrained design space
and progressively discover the design patterns.

\subsection{Design Space Experimental Setup}
\label{subsec:designspace-exp-setup}

We first describe our experimental setup for discovering the design patterns. Note that our process is generic for other tasks and future pretrained backbone models.

\paragraph{Datasets} Our process for discovering design patterns 
of PEFT 
is based on the average performances on the widely-used GLUE benchmark \citep{Wang2018GLUEAM}.  It covers a wide range of natural language understanding tasks.
First,
\textit{single-sentence tasks} include (i) Stanford Sentiment Treebank (SST-2) and 
(ii) Corpus of Linguistic Acceptability (CoLA). 
Second,
\textit{similarity and paraphrase tasks} include (i) Quora Question Pairs (QQP), 
(ii) Semantic Textual Similarity Benchmark (STS-B), 
and (iii) Microsoft Research Paraphrase Corpus (MRPC).
Third,
\textit{inference tasks} include (i) Multi-Genre Natural Language Inference (MNLI), 
(ii)  Question Natural Language Inference (QNLI), 
and (iii) Recognizing Textual Entailment (RTE). 
To compare performances,
the Matthews correlation is measured for CoLA;
the Spearman correlation is used for STS-B,
and accuracy is measured for the rest GLUE tasks.

\paragraph{Pretrained Backbone Models and Model Settings}  We use T5-base/3b \citep{raffel2020exploring} as the main pretrained backbone models for discovering design patterns via our PEFT design spaces. We use Hugging Face \footnote{\url{https://huggingface.co/docs/transformers/index}} for our implementations and follow the default settings. During the exploration, 
we set the total number of trainable parameters (in the percentage of that in the backbone model) to 0.5\% by following \citet{https://doi.org/10.48550/arxiv.2110.04366}.




\subsection{Discovering Design Patterns Using T5-base}
In this subsection,
we describe the empirical process for discovering the design patterns using T5-base (pretrained backbone model)
as the working example. Each PEFT design space (denoted as $\mathcal{S}_i$) consists of a set of models ($\mathcal{S}_i$-models) that satisfy constraints characterizing the space  with respect to layer grouping, trainable parameter allocation, tunable groups, and strategy assignment. To discover design patterns,
we start from a relatively unconstrained PEFT
design space ($\mathcal{S}_0$). Then 
we progressively refine design spaces (from $\mathcal{S}_0$ to $\mathcal{S}_{1:4}$) by
comparing overall quality of models in design spaces enforced with
different constraints (e.g., each group has the same number of layers). 
To quantify the overall quality of models in any design space $\mathcal{S}_i$
with a low-compute, low-epoch regime \citep{https://doi.org/10.48550/arxiv.2003.13678},
we randomly sample 100 models from $\mathcal{S}_i$, fine-tune with 3 epochs\footnote{We set the low epoch by observing whether it is enough for models to obtain stable performances to draw consistent conclusions (See Table~\ref{Tab:Grouping-1} in the Appendix).}, and compute the average of the GLUE average performances.

We emphasize that our goal is to demonstrate how the perspective of design spaces can help inform PEFT research, rather than to find out the ``best'' design space or method. For computational efficiency, it is beyond the scope of  this work to enumerate all possible constraints with respect to the design space components (Section \ref{sec:Components of Design Spaces}).

\subsubsection{The Initial $\mathcal{S}_0$ Design Space}
The initial relatively unconstrained design space $\mathcal{S}_0$ consists of all models without constraints  on the design space components (Section \ref{sec:Components of Design Spaces}).
Individual PEFT strategies
consist of Adapter, Prefix, BitFit, and LoRA.
One can think of
this $\mathcal{S}_0$ design space
as a set of random models ($\mathcal{S}_0$-models) with random design patterns.
Specifically, 
without grouping constraints,
each layer of the pretrained layer
has a half chance to be tuned:
if tuned, random strategies (or combinations)
with a random amount of trainable parameters
are assigned to that layer.

Before comparing more subtle design patterns 
such as how to properly assign tunable strategies
among Adapter, Prefix, BitFit, and LoRA,
we begin with exploring 
how to group layers
and how to allocate the total number of trainable parameters
to layers.

\subsubsection{The $\mathcal{S}_1$ Design Space with Additional Grouping Constraints} \label{Sec:grouping}

\begin{figure}[h!]
\centering
\includegraphics[width=0.8\textwidth]{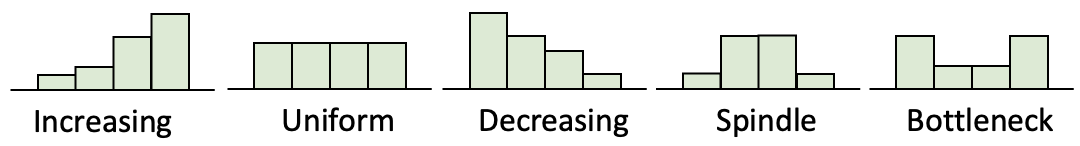}
\caption{Layer grouping patterns, where the horizontal and vertical axes represent groups ($G_1, \ldots, G_4$) and numbers of layers in groups.} \label{Fig:groups}
\end{figure}

Inspired by \citet{https://doi.org/10.48550/arxiv.2003.13678}, we also consider \textit{4} groups ($G_1, \ldots, G_4$, in the order of forward pass) in the experiments \footnote{The experimental results with 8 groups are shown in the Table ~\ref{Tab:Grouping-8-groups} in the Appendix.}.
Denote  by $N_i$ the number of layers in $G_i$.
As illustrated in Figure~\ref{Fig:groups},
we compare the following layer grouping patterns: 
(i) \textit{Increasing} ($N_{i+1} > N_i$): the number of layers in groups gradually increases; 
(ii) \textit{Uniform} ($N_{i+1} = N_i$): the number of layers in groups is the same; 
(iii) \textit{Decreasing} ($N_{i+1} < N_i$): the number of layers in groups gradually decreases; 
(iv) \textit{Spindle} (${N_1 < N_2 = N_3 > N_4}$): the numbers of layers in groups at both ends are smaller; 
and (v) \textit{Bottleneck} ($N_1 > N_2 = N_3 < N_4$): the numbers of layers in groups at both ends are bigger. 

These layer grouping patterns lead to 5 different design spaces.
Any of these 5 design spaces
consists of all models in the $\mathcal{S}_0$ design space
that satisfy one of these grouping pattern constraints.
To compare the overall model qualities of different design spaces,
we (i) randomly sample 100 models from the $\mathcal{S}_0$ design space that satisfy each grouping pattern constraint (Figure \ref{Fig:groups});
(ii) fine-tune with 3 epochs;
and (iii) compute the average performances for each design space.
We will follow this procedure as we progressively 
add new constraints later.

The averaged performances are shown in Table~\ref{Tab:Grouping-20}\footnote{The training time for the step is shown in the Table~\ref{Tab:training-time} in the Appendix.}.  We find that models from the design space with the spindle grouping pattern (Figure \ref{Fig:groups}) consistently outperform those from the other design spaces 
across all the 8 GLUE tasks. 
This may be due to the complexities of information captured in different layers of large pretrained models,
which favor information adaptation in the discovered layer grouping pattern.


\emph{From now on, we will group layers in a spindle pattern.} We refer to
$\mathcal{S}_0$ with this additional design pattern as 
the new $\mathcal{S}_1$ design space.

\begin{table}[t]
\caption{Average performances (low-compute, low-epoch regime: 100 random models, 3 tuning epochs) on the GLUE datasets using the T5-base pretrained backbone model. We compare adding different layer grouping constraints to the $\mathcal{S}_0$ design space.} \label{Tab:Grouping-20}
\begin{center}
\small
\begin{tabular}{c|cccccccc|c}
\toprule
\textbf{Layer Grouping}                             & \textbf{SST-2}              & \textbf{MNLI}                & \textbf{QNLI}                & \textbf{QQP}                 & \textbf{RTE}                & \textbf{STS-B}               & \textbf{MRPC}                & \textbf{CoLA } &\textbf{Avg}              \\ \midrule  \midrule 

 $\mathcal{S}_0$-models &76.9 & 70.1 & 72.5 & 73.3 & 63.6 & 71.7 & 73.8  & 24.3 & 65.7 \\

 \midrule

Increasing                           & 85.3          & 74.9    & 77.2          & 77.5          & 66.8        & 76.2       & 76.0          & 33.0        & 70.8 \\ 
Uniform                                      & 84.8           & 73.7        & 78.1          & 78.6         & 68.5        & 77.8         & 79.2        & 36.1         & 72.1 \\ 
Decreasing                              & 81.9           & 72.1          & 78.3          & 76.7        & 67.3          & 75.9        & 78.6       & 28.7        & 70.0 \\ 
\textbf{Spindle} & \textbf{ 86.9 } & \textbf{ 75.5 } & \textbf{ 79.8 } & \textbf{ 79.4 } & \textbf{ 69.8 } & \textbf{ 78.3 } & \textbf{ 80.1} & \textbf{ 37.3 } & \textbf{73.3} \\ 
Bottleneck         & 84.5          & 74.6          & 76.9         & 78.1           & 69.2          & 76.2           & 78.6         & 32.1    & 71.3 \\ \bottomrule    
\end{tabular} 
\end{center}
\end{table}

\begin{table}[t]
\caption{Average performances (low-compute, low-epoch regime: 100 random models, 3 tuning epochs) on the GLUE datasets using the T5-base pretrained backbone model. We compare adding different parameter allocation constraints to the $\mathcal{S}_1$ design space.} \label{Tab:Allocate}
\begin{center}
\small
\begin{tabular}{c|cccccccc|c}
\toprule
\textbf{Param Allocation} & \textbf{SST-2}                          & \textbf{MNLI}                           & \textbf{QNLI}       & \textbf{QQP}        & \textbf{RTE}        & \textbf{STS-B}      & \textbf{MRPC}       & \textbf{CoLA} & \textbf{Avg}        \\ \midrule \midrule
Increasing   & 87.2                           & \textbf{ 77.9 }                     & 79.4 & 78.7        & 71.6   & 77.6         & \textbf{ 81.4 } & 32.0  &73.2 \\
\textbf{Uniform}        & \textbf{87.8} & 77.4  & \textbf{ 80.1 } & \textbf{ 80.5 } & \textbf{ 73.9 } & \textbf{78.1} & 80.4  & 34.3  & \textbf{74.0}      \\
Decreasing       & 86.4                           & 75.8                             & 78.4          & 77.0           & 70.4           & 77.1         & 78.7       & \textbf{ 35.8 } &72.4 \\ \bottomrule
\end{tabular} 
\end{center}
\end{table}

\subsubsection{The $\mathcal{S}_2$ Design Space with Additional Parameter Constraints}

We continue to explore design patterns in trainable parameter allocation to refine the $\mathcal{S}_1$ design space. 
Denote by $n_i$ the number of trainable parameters for the $i$-th layer
of the pretrained backbone model, we compare the following design patterns: (i) \textit{Increasing} ($n_{i+1} \geq n_i$): the number of trainable parameters in every layer gradually increases (or remains the same); (ii) \textit{Uniform} (${n_{i+1} = n_i}$): the number of trainable parameters in every layer is the same; and (iii) \textit{Decreasing} ($n_{i+1} \leq n_i$): the number of trainable parameters in every layer gradually decreases (or remains the same). Following the procedure described in Section~\ref{Sec:grouping}, 
we obtain 100 models for each of these 3 new design spaces. 
Table~\ref{Tab:Allocate} reports the average performances of these 3 design spaces. 
The uniform allocation design pattern
obtains the highest 
GLUE average performance,
making this relatively simple, interpretable design pattern favorable.


\emph{We will allocate the number of trainable parameters to layers uniformly.} We refer to
$\mathcal{S}_1$ with this additional design pattern as 
the new $\mathcal{S}_2$ design space.

\begin{table}[t]

\caption{Average performances (low-compute, low-epoch regime: 100 random models, 3 tuning epochs) on the GLUE datasets using the T5-base pretrained backbone model. We compare adding different tunable group constraints to the $\mathcal{S}_2$ design space.}\label{Tab:Tunable} 
\begin{center}
\small
\begin{tabular}{c|cccccccc|c}
\toprule
\textbf{Tunable Groups} & \textbf{SST-2}      & \textbf{MNLI}       & \textbf{QNLI}        & \textbf{QQP}        & \textbf{RTE}         & \textbf{STS-B}      & \textbf{MRPC}        & \textbf{CoLA}  &\textbf{Avg}       \\ \midrule \midrule
$G_1$                      & 82.6           & 72.1           & 77.6            & 70.6           & 65.3            & 71.9           & 77.6            & 27.6   &68.2         \\
$G_2$                      & 83.3           & 72.8           & 77.5            & 72.8           & 63.6            & 72.8           & 77.5            & 27.5   &68.4         \\
$G_3$                      & 83.6           & 73.3           & 78.2            & 73.3           & 66.4            & 71.3           & 77.9            & 22.9  & 68.4           \\
$G_4$                      & 83.2           & 73.0           & 77.9            & 73.7           & 63.9            & 72.0           & 77.9            & 27.9  & 68.7           \\
$G_1$, $G_2$                  & 83.5           & 73.2           & 78.0            & 75.4           & 67.7            & 73.2           & 78.0            & 28.0   & 69.6          \\
$G_3$, $G_4$                  & 87.8           & 74.6           & 78.3            & 76.9           & 68.6            & 74.3           & 78.3            & 28.3    & 70.7         \\
$G_1$, $G_2$, $G_3$              & 86.0           &75.8           & 79.0            & 77.8           & 71.8            & 78.8           & 79.0            & 33.0   &72.6         \\
$G_2$, $G_3$, $G_4$              & 85.2           & 76.6           & 79.1            & 78.6           & 70.1            & 77.6           & 79.1            & 31.9      &72.2      \\
$\boldsymbol{G_1, G_2, G_3, G_4}$ & \textbf{88.3} & \textbf{ 77.4 } & \textbf{ 82.1 } & \textbf{ 81.5 } & \textbf{ 74.9 } & \textbf{ 79.4 } & \textbf{ 81.4 } & \textbf{34.3}  &\textbf{74.9}      \\ \bottomrule
\end{tabular}
    
\end{center}
\end{table}

\subsubsection{The $\mathcal{S}_3$ Design Space with Additional Tunable Group Constraints}

Before digging into the strategy assignment design patterns,
it is necessary to examine which groups need to be tuned.
After all,
it is only meaningful to study assigning strategies to different groups after we find out which groups need to be fine-tuned. 
As shown in Table~\ref{Tab:Tunable}, we explore various design patterns in tunable groups to further constrain the $\mathcal{S}_2$ design space. Based on the GLUE average performances, 
we find that all the groups need to be tuned to obtain the best performances.
This suggests that all the groups
of pretrained layers have
captured useful information that 
should be adapted to the downstream tasks.


\emph{We will tune all the groups.} We refer to
$\mathcal{S}_2$ with this additional design pattern as 
the new $\mathcal{S}_3$ design space.

\subsubsection{The $\mathcal{S}_4$ Design Space with Additional Strategy Constraints}
\label{subsec:s4}

Finally, we study the subtle design pattern with respect to assigning proper strategies by further constraining the derived $\mathcal{S}_3$ design space.
Specifically,
each design space consists of models
that assign a subset of \{Adapter (A), Prefix (P), BitFit (B), and LoRA (L)\} to all layers of any group $G_i$ ($i = 1, \ldots, 4$).
We begin by 
adding different $G_1$ strategy assignment constraints 
to the $\mathcal{S}_3$ space.
Following the same pattern discovery procedure (Section~\ref{Sec:grouping}),
we discover strategy assignment patterns for $G_1$.
Then we progressively
add $G_i$ ($i>1$) strategy assignment constraints 
together with the discovered strategy assignment patterns for all $G_j$ ($j = 1, \ldots, i-1$)
to the $\mathcal{S}_3$ space.
Due to space limit, 
we present results of this process
in the Appendix ($G_1$ in Table~\ref{Tab:$G_1$}, $G_2$ Table~\ref{Tab:$G_2$}, $G_3$ in Table~\ref{Tab:$G_3$}, and $G_4$ in Table~\ref{Tab:$G_4$}),
which suggests strategy assignment of
$G_1$-(A, L)  -- $G_2$-(A, P) -- $G_3$-(A, P, B)  -- $G_4$-(P, B, L)
for the T5-base pretrained backbone model.


\emph{We will assign the discovered proper tuning strategies to groups.} We refer to
$\mathcal{S}_3$ with this additional design pattern as 
the new $\mathcal{S}_4$ design space, which consists of the final  $\mathcal{S}_4$-model.

\subsection{Discovering Design Patterns Using T5-3b}
\label{subsec:s4-3b}

We then repeat the above process on T5-3b to examine if the design patterns we discovered using smaller models (T5-base) still apply when we use larger models. The results are shown in Table~\ref{Tab:Grouping-3-3b} (layer grouping), Table~\ref{Tab:Allocate-3b} (trainable parameter allocation),  Table~\ref{Tab:Tunable-3b} (tunable groups) and Table~\ref{Tab:g-3b} (strategy assignment) in the Appendix. We observe that the design patterns still apply when larger models like T5-3b are used: 
(i) grouping layers in a spindle pattern (Table~\ref{Tab:Grouping-3-3b}), (ii) uniformly allocating the number of trainable parameters to layers (Table~\ref{Tab:Allocate-3b}), (iii) tuning all the groups (Table~\ref{Tab:Tunable-3b}), and (iv) tuning different groups with proper strategies (Table~\ref{Tab:g-3b}).
For T5-3b, the discovered proper strategy assignment is
$G_1$-(P, L)  -- $G_2$-(A, L) -- $G_3$-(P, B, L)  -- $G_4$-(A, P, B).
We refer to the final design space as $\mathcal{S}_4$-3b and the final model in this space as $\mathcal{S}_4$-3b-model.

\begin{table}[t]
 \caption{Performances of different tuning methods on the GLUE datasets using the T5-base (upper part) and T5-3b (lower part) pretrained backbone models, respectively. The results are averaged over 20 random runs (with standard deviations as subscripts). The $\mathcal{S}_4$-model and the $\mathcal{S}_4$-3b-model perform significantly better than the second-best PEFT methods in all the eight datasets at the significance level $p<0.05(*)$ or even $p<0.01(**)$. } \label{Tab:ALL}
 \begin{center}
\scriptsize
\begin{tabular}{c|cccccccc|c} \toprule
\textbf{Method}     & \multicolumn{1}{l}{\textbf{SST-2}} & \multicolumn{1}{l}{\textbf{MNLI}} & \multicolumn{1}{l}{\textbf{QNLI}} & \multicolumn{1}{l}{\textbf{QQP}} & \multicolumn{1}{l}{\textbf{RTE}} & \multicolumn{1}{l}{\textbf{STS-B}} & \multicolumn{1}{l}{\textbf{MRPC}} & \multicolumn{1}{l}{\textbf{CoLA}} & \multicolumn{1}{|l}{\textbf{Average}} \\ \midrule \midrule
full              & 95.2          & 87.1          & 93.7 & 89.4 & 80.1          & 89.4 & 90.7          & 51.1          & 84.5          \\ \midrule
\underline{Adapter}           & 94.6          & 85.5          & 89.8          & 86.7          & 75.3          & 86.7          & 89.1          & 59.2          & 83.3          \\
Prefix            & 94.0            & 81.6          & 87.8          & 83.4          & 64.3          & 83.1          & 84.8          & 34.0            & 76.6          \\
BitFit            & 94.4          & 84.5          & 90.6          & 88.3          & 74.3          & 86.6          & 90.1          & 57.7          & 83.3          \\
LoRA              & 94.8          & 84.7          & 91.6          & 88.5          & 75.8          & 86.3          & 88.7          & 51.5          & 82.7          \\ 
\textbf{$\mathcal{S}_4$-model} & \textbf{$\mathbf{95.5}_{1.7}^{**}$} & \textbf{$\mathbf{87.6}_{1.0}^{**}$} & \textbf{$\mathbf{92.7}_{1.1}^{**}$}          & \textbf{$\mathbf{88.8}_{1.0}^{**}$}          & \textbf{$\mathbf{80.4}_{2.3}^{*}$} & \textbf{$\mathbf{87.4}_{2.0}^{*}$}          & \textbf{$\mathbf{91.2}_{2.4}^{**}$} & \textbf{$\mathbf{62.2}_{3.2}^{*}$} & \textbf{85.7} 

\\ \midrule \midrule

full             & 97.4                      & 91.4                              & 96.3                              & 89.7                    & 91.1                             & 90.6                               & 92.5                     & 67.1                              & 89.5       \\ \midrule
Adapter            & 96.3                               & 89.9                              & 94.7                              & 87.8                             & 83.4                             & 90                                 & 89.7                              & 65.2                              & 87.1                     \\
Prefix           & 96.3                               & 82.8                              & 88.9                              & 85.5                             & 78.3                             & 83.5                               & 85.4                              & 42.7                              & 80.4                   \\
BitFit              & 95.8                               & 89.5                              & 93.5                              & 88.5                             & 86.2                             & 90.7                               & 88.6                              & 64.2                              & 87.1                 \\
\underline{LoRA}             & 96.2                               & 90.6                              & 94.9                              & 89.1                             & 91.2                             & 91.1                               & 91.1                              & 67.4                              & 88.9                  \\
\textbf{$\mathcal{S}_4$-3b-model}  & \textbf{$\mathbf{97.2}_{1.8}^{**}$}                               & \textbf{$\mathbf{91.6}_{1.2}^{**}$}                     & \textbf{$\mathbf{96.6}_{1.0}^{**}$}                     & \textbf{$\mathbf{89.5}_{1.5}^{**}$}                             & \textbf{$\mathbf{91.5}_{2.8}^{*}$}                    & \textbf{$\mathbf{91.5}_{2.5}^{*}$}                      & \textbf{$\mathbf{91.9}_{2.0}^{*}$}                              & \textbf{$\mathbf{69.7}_{3.4}^{*}$}                     & \textbf{89.9}  
\\ \bottomrule                          
\end{tabular}
 \end{center}
\end{table}

\begin{table}[t]
 \caption{Performances of different tuning methods on GLUE datasets using the RoBERTa-base (upper part) and RoBERTa-large (lower part) pretrained backbone models. The results are averaged over 20 random runs (with standard deviations as subscripts). Here we also include two baselines: (i) \textit{$\mathcal{S}_0$-model, where all the designs are randomly selected for RoBERTa as in the $\mathcal{S}_0$ design space; (ii) \textit{$\mathcal{S}_3$-model}, where strategies are randomly assigned to different RoBERTa layer groups as in the $\mathcal{S}_3$ design space. The $\mathcal{S}_4$-model and $\mathcal{S}_4$-3b-model perform significantly better than the second-best PEFT methods in all the eight datasets at the significance level $p<0.05(*)$ or even $p<0.01(**)$.}} \label{Tab:ALL-roberta}
\begin{center}

\scriptsize
\begin{tabular}{c|cccccccc|c} \toprule
\textbf{Method}     & \multicolumn{1}{l}{\textbf{SST-2}} & \multicolumn{1}{l}{\textbf{MNLI}} & \multicolumn{1}{l}{\textbf{QNLI}} & \multicolumn{1}{l}{\textbf{QQP}} & \multicolumn{1}{l}{\textbf{RTE}} & \multicolumn{1}{l}{\textbf{STS-B}} & \multicolumn{1}{l}{\textbf{MRPC}} & \multicolumn{1}{l}{\textbf{CoLA}} & \multicolumn{1}{l}{\textbf{Average}} \\ \midrule \midrule
full                           & 94.8                               & 87.6                              & 92.8                              & 91.9                    & 80.8                    & 90.3                               & 90.2                     & 63.6                     & 86.5                 \\ \midrule
Adapter                        & 94.2                               & 87.1                              & 93.1                              & 90.2                             & 71.5                             & 89.7                               & 88.5                              & 60.8                              & 84.4                 \\
Prefix                         & 94.0                               & 86.8                              & 91.3                              & 90.5                             & 74.5                             & 90.3                               & 88.2                              & 61.5                              & 84.6                 \\

BitFit                         & 93.7                               & 84.8                              & 91.3                              & 84.5                             & 77.8                             & \textbf{90.8}                      & 90.0                              & 61.8                              & 84.3                 \\
\underline{LoRA}                           & \textbf{94.9}                      & 87.5                              & 93.1                              & 90.8                             & 83.1                             & 90.0                               & 89.6                              & 62.6                              & 86.4                 \\
$\mathcal{S}_0$-model                       & 94.2                               & 95.3                              & 90.4                              & 90.6                             & 75.6                             & 89.6                               & 88.0                              & 60.9                              & 85.6                 \\
$\mathcal{S}_3$-model                       & 94.3                               & 87.2                              & 92.8                              & 91.0                             & 81.8                             & 90.3                               & 89.2                              & 63.2                              & 86.2                 \\
\textbf{$\mathcal{S}_4$-model}              & $94.8_{1.6}$                               & \textbf{$\mathbf{87.8}_{0.8}^{**}$}                     & \textbf{$\mathbf{93.4}_{1.3}^{**}$}                     & $\mathbf{91.6}_{1.2}^{*}$                             & \textbf{$\mathbf{85.8}_{1.8}^{**}$}                    & $90.4_{2.0}^{*}$                               & $\mathbf{90.0}_{1.8}^{**}$                              & $\mathbf{63.2}_{3.5}^{*}$                              & \textbf{87.1}             
\\  \midrule \midrule

full                           & 96.4                               & 90.2                              & 94.7                              & 92.2                    & 86.6                             & 92.4                      & 90.9                              & 68.0                              & 88.9                 \\ \midrule
Adapter                        & 96.6                               & 90.5                              & 94.8                              & 91.7                             & 80.1                             & 92.1                               & 90.9                              & 67.8                              & 88.1                 \\
Prefix                         & 95.7                               & 87.6                              & 92.1                              & 88.7                             & 82.3                             & 89.6                               & 87.4                              & 62.8                              & 85.7                 \\

BitFit                         & 96.1                               & 88.0                                & 93.4                              & 90.2                             & 86.2                             & 90.9                               & \textbf{92.7}                     & 64.2                              & 87.7                 \\
\underline{LoRA}                           & 96.2                               & 90.6                              & 94.7                              & 91.6                             & \textbf{87.4}                    & 92.0                               & 89.7                              & 68.2                              & 88.8                 \\
$\mathcal{S}_0$-model                       & 95.5                               & 86.5                              & 92.3                              & 89.8                             & 84.6                             & 89.2                               & 86.3                              & 61.2                              & 85.6                 \\
$\mathcal{S}_3$-model                       & 96.3                               & 89.4                              & 93.8                              & 90.2                             & 85.9                             & 90.8                               & 90.9                              & 63.4                              & 87.6                 \\
\textbf{$\mathcal{S}_4$-3b-model}              & \textbf{$\mathbf{96.6}_{1.3}^{**}$}                      & \textbf{$\mathbf{90.8}_{1.1}^{*}$}                     & \textbf{$\mathbf{95.1}_{0.8}^{**}$}                     & $\mathbf{92.0}_{1.2}^{**}$                             & $87.2_{2.8}$                             & $\mathbf{92.3}_{2.2}^{*}$                               & $91.8_{1.8}^{**}$                             & \textbf{$\mathbf{68.4}_{3.2}^{*}$}                     & \textbf{89.3}              
\\

\bottomrule                          
\end{tabular}
\end{center}
\end{table}

\section{Evaluation}\label{Sec:Experiment}

The $\mathcal{S}_4$-model (Section \ref{subsec:s4}) and $\mathcal{S}_4$-3b-model (Section \ref{subsec:s4-3b})
adopt all the design patterns that have been discovered by using T5-base and T5-3b, respectively.
As a result, they are both new methods of PEFT.
We will evaluate their effectiveness when applied to different pretrained backbone models and different NLP tasks.

\subsection{Experimental Setup}

\paragraph{Datasets}
Besides the GLUE datasets \citep{Wang2018GLUEAM} (Section \ref{subsec:designspace-exp-setup}), we further evaluate our methods on two generation tasks used by \citet{https://doi.org/10.48550/arxiv.2110.04366}: (i) \textit{Abstractive Summarization} using XSum \citep{narayan-etal-2018-dont}, and (ii)  \textit{Machine Translation} using the WMT 2016 en-ro dataset \citep{bojar-etal-2016-findings}. We report ROUGE scores \citep{lin-2004-rouge} on the XSum test set, and BLEU scores \citep{Papineni2002BleuAM} on the en-ro test set.

\paragraph{Models and Model Settings}
We mainly compare our methods with the following baselines: (i) \textbf{Full Fine-tuning} (full): it fine-tunes all the model parameters in the pretrained models;
(ii) \textbf{Adapter} \citep{pmlr-v97-houlsby19a}: it adds adapter modules to each transformer layer; 
(iii) \textbf{Prefix} \citep{li2021Prefixtuning}: it optimizes a set of small continuous vectors prepended to transformer layers;  
(iv) \textbf{BitFit} \citep{https://doi.org/10.48550/arxiv.2106.10199}: it only updates the bias terms in pretrained models; 
(v) \textbf{LoRA} \citep{https://doi.org/10.48550/arxiv.2106.09685}: it decomposes the attention weight into low-rank matrices to reduce the number of trainable parameters. Besides T5 \citep{raffel2020exploring}, we additionally apply our methods to other backbone models including RoBERTa-base/large \citep{liu2019roberta} and BART-base/large \citep{lewis2019bart}. We use the default settings.  
We set 
the total number of trainable parameters (in the percentage of that in the backbone model) by following \citet{https://doi.org/10.48550/arxiv.2110.04366}. Specifically, this value is set to 0.5\% for  Adapter, Prefix,  LoRA,  and our  methods, and 0.1\% for  BitFit. 

For all the experiments, we followed \citet{liu2019roberta} to set the linear decay scheduler with a warmup ratio of 0.06 for training. The batch size was 128 for base models and 64 for large models. The maximum learning rate was $5e-5$ and the maximum number of training epochs was set to be either $5$ or $10$. All the experiments were performed using 8 A100 GPUs.

\begin{wraptable}[22]{r}{0.55\textwidth}
\caption{Performances of different tuning methods on generation tasks (XSUM and en-ro) using the BART-base (upper part) and BART-large (lower part) pretrained backbone models.} \label{Tab:ALL-bart}
\begin{center}
\small
\begin{tabular}{c|cc} \toprule
\textbf{Method}     & \multicolumn{1}{l}{\textbf{XSUM(R-1/2/L)}} & \multicolumn{1}{l}{\textbf{en-ro (BLEU)}}\\ \midrule \midrule
full              & 40.5/19.2/34.8          & 34.5                               \\ \midrule 
Adapter           & 37.7/17.9/33.1          & 33.3                               \\
Prefix            & 38.2/18.4/32.4          & 33.8                               \\
BitFit            & 37.2/17.5/31.4          & 33.2                               \\
LoRA              & 38.9/18.6/33.5          & 33.6                               \\
PA                & 39.3/18.7/33.8          & 33.8                               \\
\textbf{$\mathcal{S}_4$-model} & \textbf{40.2/19.3/34.2} & \textbf{34.1}            
\\  \midrule \midrule 

full              & 45.1/22.3/37.2          & 37.9                               \\\midrule 
Adapter           & 43.8/20.8/35.7          & 35.3                               \\
Prefix            & 43.4/20.4/35.5          & 35.6                               \\
BitFit            & 42.8/18.7/33.2          & 35.2                               \\
LoRA              & 42.9/19.4/34.8          & 35.8                               \\
PA                & 43.9/20.6/35.6          & 36.4                               \\
\textbf{$\mathcal{S}_4$-3b-model} & \textbf{44.3/21.7/36.8} & \textbf{37.2}              
\\ 
\bottomrule                          
\end{tabular}
\end{center}
\end{wraptable}


\subsection{Effectiveness on GLUE with T5 Backbones}

With our discovered design patterns, we fine-tune T5-base ($\mathcal{S}_4$-model) and T5-3b ($\mathcal{S}_4$-3b-model) on GLUE and compare them with all the baseline methods. The results are shown in Table~\ref{Tab:ALL}, where the key measure is the GLUE average performance (last column).
We find that our $\mathcal{S}_4$-model and  $\mathcal{S}_4$-3b-model consistently outperform  the investigated methods in the key measure. By tuning only $0.5\%$ parameters, our methods even outperform the full fine-tuning baseline where all the parameters are tuned, indicating the effectiveness of our discovered PEFT design patterns.

\subsection{General Effectiveness on GLUE with RoBERTa Backbones}
\label{subsec:effective-roberta}

We directly apply the 
$\mathcal{S}_4$-model and $\mathcal{S}_4$-3b-model (adopting design patterns discovered using T5-base and T5-3b) to fine-tune the RoBERTa-base and RoBERTa-large pretrained backbone models (with no extra discovery process), respectively.
We keep all the other settings the same and evaluate them on GLUE datasets. 
We also compare with variant methods randomly sampled from two design spaces: (i) \textit{$\mathcal{S}_0$-model}, where all the designs are randomly selected for RoBERTa as in $\mathcal{S}_0$; (ii) \textit{$\mathcal{S}_3$-model}, where strategies are randomly assigned to different RoBERTa layer groups as in $\mathcal{S}_3$.
Table~\ref{Tab:ALL-roberta} shows that (i) the design patterns (adopted by $\mathcal{S}_4$-model and $\mathcal{S}_4$-3b-model) discovered using T5 models are applicable to the RoBERTa backbone models
and outperform the investigated methods in GLUE average performances
with no extra discovery process;
(ii) improved performances from $\mathcal{S}_0$-models, $\mathcal{S}_3$-models, to $\mathcal{S}_4$-(3b)-models support adding more constraints in the pattern discovery process (Section \ref{Sec:Searching}).

\subsection{General Effectiveness on Generation Tasks with BART Backbones}

Like in Section \ref{subsec:effective-roberta},
we further directly apply the 
$\mathcal{S}_4$-model and $\mathcal{S}_4$-3b-model (adopting design patterns discovered using T5-base and T5-3b) to fine-tune the BART-base and BART-large pretrained backbone models (without additional discovery process.), respectively.
We evaluate the models on two generation tasks: summarization (XSUM) and machine translation (en-ro) following \citet{https://doi.org/10.48550/arxiv.2110.04366}. 
We also compare with
PA (parallel adapter) using the same number of trainable parameters \citep{https://doi.org/10.48550/arxiv.2110.04366}. 
Table~\ref{Tab:ALL-bart} shows that 
our methods, although adopting design patterns discovered from classification tasks using T5,
still outperform investigated
PEFT strategies
on generation tasks with different BART backbones.


\section{Conclusion}

PEFT
adapts knowledge in pretrained models to down-stream tasks in a more parameter-efficient fashion.
Instead of focusing on designing another strategy in the first place,
we introduced PEFT design spaces. 
We empirically discovered several design patterns in PEFT.
These design patterns led to
new PEFT methods.
Experiments showed that these methods consistently outperform investigated PEFT strategies across different backbone models and different tasks in natural language processing.

\bibliography{draft}

\begin{thebibliography}{49}
\providecommand{\natexlab}[1]{#1}
\providecommand{\url}[1]{\texttt{#1}}
\expandafter\ifx\csname urlstyle\endcsname\relax
  \providecommand{\doi}[1]{doi: #1}\else
  \providecommand{\doi}{doi: \begingroup \urlstyle{rm}\Url}\fi

\bibitem[Devlin et~al.(2019)Devlin, Chang, Lee, and Toutanova]{devlin2018bert}
Jacob Devlin, Ming-Wei Chang, Kenton Lee, and Kristina Toutanova.
\newblock Bert: Pre-training of deep bidirectional transformers for language
  understanding.
\newblock In \emph{NAACL-HLT}, 2019.

\bibitem[Liu et~al.(2019)Liu, Ott, Goyal, Du, Joshi, Chen, Levy, Lewis,
  Zettlemoyer, and Stoyanov]{liu2019roberta}
Yinhan Liu, Myle Ott, Naman Goyal, Jingfei Du, Mandar Joshi, Danqi Chen, Omer
  Levy, Mike Lewis, Luke Zettlemoyer, and Veselin Stoyanov.
\newblock Roberta: A robustly optimized bert pretraining approach.
\newblock \emph{arXiv preprint arXiv:1907.11692}, 2019.

\bibitem[Yang et~al.(2019)Yang, Dai, Yang, Carbonell, Salakhutdinov, and
  Le]{yang2019xlnet}
Zhilin Yang, Zihang Dai, Yiming Yang, Jaime Carbonell, Russ~R Salakhutdinov,
  and Quoc~V Le.
\newblock Xlnet: Generalized autoregressive pretraining for language
  understanding.
\newblock In \emph{Advances in neural information processing systems}, pages
  5754--5764, 2019.

\bibitem[Joshi et~al.(2019)Joshi, Chen, Liu, Weld, Zettlemoyer, and
  Levy]{joshi2019spanbert}
Mandar Joshi, Danqi Chen, Yinhan Liu, Daniel~S. Weld, Luke Zettlemoyer, and
  Omer Levy.
\newblock Spanbert: Improving pre-training by representing and predicting
  spans.
\newblock \emph{Transactions of the Association for Computational Linguistics},
  8:\penalty0 64--77, 2019.

\bibitem[Sun et~al.(2019)Sun, Wang, Li, Feng, Chen, Zhang, Tian, Zhu, Tian, and
  Wu]{sun2019ernie}
Yu~Sun, Shuohuan Wang, Yukun Li, Shikun Feng, Xuyi Chen, Han Zhang, Xin Tian,
  Danxiang Zhu, Hao Tian, and Hua Wu.
\newblock Ernie: Enhanced representation through knowledge integration.
\newblock \emph{arXiv preprint arXiv:1904.09223}, 2019.

\bibitem[Clark et~al.(2019)Clark, Luong, Le, and Manning]{clark2019electra}
Kevin Clark, Minh-Thang Luong, Quoc~V Le, and Christopher~D Manning.
\newblock Electra: Pre-training text encoders as discriminators rather than
  generators.
\newblock In \emph{International Conference on Learning Representations}, 2019.

\bibitem[Lewis et~al.(2020{\natexlab{a}})Lewis, Liu, Goyal, Ghazvininejad,
  Mohamed, Levy, Stoyanov, and Zettlemoyer]{lewis2019bart}
Mike Lewis, Yinhan Liu, Naman Goyal, Marjan Ghazvininejad, Abdelrahman Mohamed,
  Omer Levy, Ves Stoyanov, and Luke Zettlemoyer.
\newblock Bart: Denoising sequence-to-sequence pre-training for natural
  language generation, translation, and comprehension.
\newblock \emph{SCL}, 2020{\natexlab{a}}.

\bibitem[Bao et~al.(2020)Bao, Dong, Wei, Wang, Yang, Liu, Wang, Piao, Gao,
  Zhou, et~al.]{bao2020unilmv2}
Hangbo Bao, Li~Dong, Furu Wei, Wenhui Wang, Nan Yang, Xiaodong Liu, Yu~Wang,
  Songhao Piao, Jianfeng Gao, Ming Zhou, et~al.
\newblock Unilmv2: Pseudo-masked language models for unified language model
  pre-training.
\newblock \emph{arXiv preprint arXiv:2002.12804}, 2020.

\bibitem[He et~al.(2020)He, Liu, Gao, and Chen]{he2020deberta}
Pengcheng He, Xiaodong Liu, Jianfeng Gao, and Weizhu Chen.
\newblock Deberta: Decoding-enhanced bert with disentangled attention.
\newblock \emph{arXiv preprint arXiv:2006.03654}, 2020.

\bibitem[Raffel et~al.(2020)Raffel, Shazeer, Roberts, Lee, Narang, Matena,
  Zhou, Li, and Liu]{raffel2020exploring}
Colin Raffel, Noam Shazeer, Adam Roberts, Katherine Lee, Sharan Narang, Michael
  Matena, Yanqi Zhou, Wei Li, and Peter~J. Liu.
\newblock Exploring the limits of transfer learning with a unified text-to-text
  transformer, 2020.

\bibitem[Ziems et~al.(2022)Ziems, Chen, Harris, Anderson, and
  Yang]{ziems-etal-2022-value}
Caleb Ziems, Jiaao Chen, Camille Harris, Jessica Anderson, and Diyi Yang.
\newblock {VALUE}: {U}nderstanding dialect disparity in {NLU}.
\newblock In \emph{Proceedings of the 60th Annual Meeting of the Association
  for Computational Linguistics (Volume 1: Long Papers)}, pages 3701--3720,
  Dublin, Ireland, May 2022. Association for Computational Linguistics.
\newblock \doi{10.18653/v1/2022.acl-long.258}.
\newblock URL \url{https://aclanthology.org/2022.acl-long.258}.

\bibitem[Brown et~al.(2020)Brown, Mann, Ryder, Subbiah, Kaplan, Dhariwal,
  Neelakantan, Shyam, Sastry, Askell, Agarwal, Herbert-Voss, Krueger, Henighan,
  Child, Ramesh, Ziegler, Wu, Winter, Hesse, Chen, Sigler, Litwin, Gray, Chess,
  Clark, Berner, McCandlish, Radford, Sutskever, and Amodei]{brown2020language}
Tom~B. Brown, Benjamin Mann, Nick Ryder, Melanie Subbiah, Jared Kaplan,
  Prafulla Dhariwal, Arvind Neelakantan, Pranav Shyam, Girish Sastry, Amanda
  Askell, Sandhini Agarwal, Ariel Herbert-Voss, Gretchen Krueger, Tom Henighan,
  Rewon Child, Aditya Ramesh, Daniel~M. Ziegler, Jeffrey Wu, Clemens Winter,
  Christopher Hesse, Mark Chen, Eric Sigler, Mateusz Litwin, Scott Gray,
  Benjamin Chess, Jack Clark, Christopher Berner, Sam McCandlish, Alec Radford,
  Ilya Sutskever, and Dario Amodei.
\newblock Language models are few-shot learners, 2020.

\bibitem[Houlsby et~al.(2019{\natexlab{a}})Houlsby, Giurgiu, Jastrzebski,
  Morrone, De~Laroussilhe, Gesmundo, Attariyan, and Gelly]{pmlr-v97-houlsby19a}
Neil Houlsby, Andrei Giurgiu, Stanislaw Jastrzebski, Bruna Morrone, Quentin
  De~Laroussilhe, Andrea Gesmundo, Mona Attariyan, and Sylvain Gelly.
\newblock Parameter-efficient transfer learning for {NLP}.
\newblock In Kamalika Chaudhuri and Ruslan Salakhutdinov, editors,
  \emph{Proceedings of the 36th International Conference on Machine Learning},
  volume~97 of \emph{Proceedings of Machine Learning Research}, pages
  2790--2799. PMLR, 09--15 Jun 2019{\natexlab{a}}.
\newblock URL \url{http://proceedings.mlr.press/v97/houlsby19a.html}.

\bibitem[Pfeiffer et~al.(2021)Pfeiffer, Kamath, R{\"u}ckl{\'e}, Cho, and
  Gurevych]{pfeiffer-etal-2021-adapterfusion}
Jonas Pfeiffer, Aishwarya Kamath, Andreas R{\"u}ckl{\'e}, Kyunghyun Cho, and
  Iryna Gurevych.
\newblock {A}dapter{F}usion: Non-destructive task composition for transfer
  learning.
\newblock In \emph{Proceedings of the 16th Conference of the European Chapter
  of the Association for Computational Linguistics: Main Volume}, pages
  487--503, Online, April 2021. Association for Computational Linguistics.
\newblock URL \url{https://www.aclweb.org/anthology/2021.eacl-main.39}.

\bibitem[Li and Liang(2021)]{li2021Prefixtuning}
Xiang~Lisa Li and Percy Liang.
\newblock Prefix-tuning: Optimizing continuous prompts for generation, 2021.

\bibitem[Lester et~al.(2021{\natexlab{a}})Lester, Al-Rfou, and
  Constant]{lester2021power}
Brian Lester, Rami Al-Rfou, and Noah Constant.
\newblock The power of scale for parameter-efficient prompt tuning,
  2021{\natexlab{a}}.

\bibitem[Schick and Sch{\"u}tze(2021)]{schick-schutze-2021-exploiting}
Timo Schick and Hinrich Sch{\"u}tze.
\newblock Exploiting cloze-questions for few-shot text classification and
  natural language inference.
\newblock In \emph{Proceedings of the 16th Conference of the European Chapter
  of the Association for Computational Linguistics: Main Volume}, pages
  255--269, Online, April 2021. Association for Computational Linguistics.
\newblock \doi{10.18653/v1/2021.eacl-main.20}.
\newblock URL \url{https://aclanthology.org/2021.eacl-main.20}.

\bibitem[Lester et~al.(2021{\natexlab{b}})Lester, Al-Rfou, and
  Constant]{lester-etal-2021-power}
Brian Lester, Rami Al-Rfou, and Noah Constant.
\newblock The power of scale for parameter-efficient prompt tuning.
\newblock In \emph{Proceedings of the 2021 Conference on Empirical Methods in
  Natural Language Processing}, pages 3045--3059, Online and Punta Cana,
  Dominican Republic, November 2021{\natexlab{b}}. Association for
  Computational Linguistics.
\newblock \doi{10.18653/v1/2021.emnlp-main.243}.
\newblock URL \url{https://aclanthology.org/2021.emnlp-main.243}.

\bibitem[Zaken et~al.(2021)Zaken, Ravfogel, and
  Goldberg]{https://doi.org/10.48550/arxiv.2106.10199}
Elad~Ben Zaken, Shauli Ravfogel, and Yoav Goldberg.
\newblock Bitfit: Simple parameter-efficient fine-tuning for transformer-based
  masked language-models, 2021.
\newblock URL \url{https://arxiv.org/abs/2106.10199}.

\bibitem[Hu et~al.(2021)Hu, Shen, Wallis, Allen-Zhu, Li, Wang, Wang, and
  Chen]{https://doi.org/10.48550/arxiv.2106.09685}
Edward~J. Hu, Yelong Shen, Phillip Wallis, Zeyuan Allen-Zhu, Yuanzhi Li, Shean
  Wang, Lu~Wang, and Weizhu Chen.
\newblock Lora: Low-rank adaptation of large language models, 2021.
\newblock URL \url{https://arxiv.org/abs/2106.09685}.

\bibitem[He et~al.(2022)He, Zhou, Ma, Berg-Kirkpatrick, and
  Neubig]{https://doi.org/10.48550/arxiv.2110.04366}
Junxian He, Chunting Zhou, Xuezhe Ma, Taylor Berg-Kirkpatrick, and Graham
  Neubig.
\newblock Towards a unified view of parameter-efficient transfer learning.
\newblock In \emph{International Conference on Learning Representations}, 2022.

\bibitem[Mao et~al.(2022)Mao, Mathias, Hou, Almahairi, Ma, Han, Yih, and
  Khabsa]{mao-etal-2022-unipelt}
Yuning Mao, Lambert Mathias, Rui Hou, Amjad Almahairi, Hao Ma, Jiawei Han,
  Scott Yih, and Madian Khabsa.
\newblock {U}ni{PELT}: A unified framework for parameter-efficient language
  model tuning.
\newblock In \emph{Proceedings of the 60th Annual Meeting of the Association
  for Computational Linguistics (Volume 1: Long Papers)}, pages 6253--6264,
  Dublin, Ireland, May 2022. Association for Computational Linguistics.
\newblock \doi{10.18653/v1/2022.acl-long.433}.
\newblock URL \url{https://aclanthology.org/2022.acl-long.433}.

\bibitem[Lewis et~al.(2020{\natexlab{b}})Lewis, Liu, Goyal, Ghazvininejad,
  Mohamed, Levy, Stoyanov, and Zettlemoyer]{lewis-etal-2020-bart}
Mike Lewis, Yinhan Liu, Naman Goyal, Marjan Ghazvininejad, Abdelrahman Mohamed,
  Omer Levy, Veselin Stoyanov, and Luke Zettlemoyer.
\newblock {BART}: Denoising sequence-to-sequence pre-training for natural
  language generation, translation, and comprehension.
\newblock In \emph{Proceedings of the 58th Annual Meeting of the Association
  for Computational Linguistics}, pages 7871--7880, Online, July
  2020{\natexlab{b}}. Association for Computational Linguistics.
\newblock \doi{10.18653/v1/2020.acl-main.703}.
\newblock URL \url{https://www.aclweb.org/anthology/2020.acl-main.703}.

\bibitem[Radosavovic et~al.(2019)Radosavovic, Johnson, Xie, Lo, and
  Dollár]{https://doi.org/10.48550/arxiv.1905.13214}
Ilija Radosavovic, Justin Johnson, Saining Xie, Wan-Yen Lo, and Piotr Dollár.
\newblock On network design spaces for visual recognition, 2019.
\newblock URL \url{https://arxiv.org/abs/1905.13214}.

\bibitem[Radosavovic et~al.(2020)Radosavovic, Kosaraju, Girshick, He, and
  Dollár]{https://doi.org/10.48550/arxiv.2003.13678}
Ilija Radosavovic, Raj~Prateek Kosaraju, Ross Girshick, Kaiming He, and Piotr
  Dollár.
\newblock Designing network design spaces, 2020.
\newblock URL \url{https://arxiv.org/abs/2003.13678}.

\bibitem[You et~al.(2020)You, Ying, and
  Leskovec]{https://doi.org/10.48550/arxiv.2011.08843}
Jiaxuan You, Rex Ying, and Jure Leskovec.
\newblock Design space for graph neural networks, 2020.
\newblock URL \url{https://arxiv.org/abs/2011.08843}.

\bibitem[Houlsby et~al.(2019{\natexlab{b}})Houlsby, Giurgiu, Jastrzebski,
  Morrone, De~Laroussilhe, Gesmundo, Attariyan, and
  Gelly]{houlsby2019parameter}
Neil Houlsby, Andrei Giurgiu, Stanislaw Jastrzebski, Bruna Morrone, Quentin
  De~Laroussilhe, Andrea Gesmundo, Mona Attariyan, and Sylvain Gelly.
\newblock Parameter-efficient transfer learning for nlp.
\newblock In \emph{International Conference on Machine Learning}, pages
  2790--2799. PMLR, 2019{\natexlab{b}}.

\bibitem[Stickland and Murray(2019)]{stickland2019bert}
Asa~Cooper Stickland and Iain Murray.
\newblock Bert and pals: Projected attention layers for efficient adaptation in
  multi-task learning.
\newblock In \emph{International Conference on Machine Learning}, pages
  5986--5995. PMLR, 2019.

\bibitem[Pfeiffer et~al.(2020)Pfeiffer, Kamath, R{\"u}ckl{\'e}, Cho, and
  Gurevych]{pfeiffer2020adapterfusion}
Jonas Pfeiffer, Aishwarya Kamath, Andreas R{\"u}ckl{\'e}, Kyunghyun Cho, and
  Iryna Gurevych.
\newblock Adapterfusion: Non-destructive task composition for transfer
  learning.
\newblock \emph{arXiv preprint arXiv:2005.00247}, 2020.

\bibitem[Rebuffi et~al.(2017)Rebuffi, Bilen, and Vedaldi]{rebuffi2017learning}
Sylvestre-Alvise Rebuffi, Hakan Bilen, and Andrea Vedaldi.
\newblock Learning multiple visual domains with residual adapters.
\newblock \emph{arXiv preprint arXiv:1705.08045}, 2017.

\bibitem[Lin et~al.(2020)Lin, Madotto, and Fung]{lin2020exploring}
Zhaojiang Lin, Andrea Madotto, and Pascale Fung.
\newblock Exploring versatile generative language model via parameter-efficient
  transfer learning.
\newblock \emph{arXiv preprint arXiv:2004.03829}, 2020.

\bibitem[Zhao et~al.(2020)Zhao, Lin, Mi, Jaggi, and
  Sch{\"u}tze]{zhao2020masking}
Mengjie Zhao, Tao Lin, Fei Mi, Martin Jaggi, and Hinrich Sch{\"u}tze.
\newblock Masking as an efficient alternative to finetuning for pretrained
  language models.
\newblock \emph{arXiv preprint arXiv:2004.12406}, 2020.

\bibitem[Guo et~al.(2020)Guo, Rush, and Kim]{guo2020parameter}
Demi Guo, Alexander~M Rush, and Yoon Kim.
\newblock Parameter-efficient transfer learning with diff pruning.
\newblock \emph{arXiv preprint arXiv:2012.07463}, 2020.

\bibitem[Mallya et~al.(2018)Mallya, Davis, and Lazebnik]{mallya2018piggyback}
Arun Mallya, Dillon Davis, and Svetlana Lazebnik.
\newblock Piggyback: Adapting a single network to multiple tasks by learning to
  mask weights.
\newblock In \emph{Proceedings of the European Conference on Computer Vision
  (ECCV)}, pages 67--82, 2018.

\bibitem[Radiya-Dixit and Wang(2020)]{radiya2020fine}
Evani Radiya-Dixit and Xin Wang.
\newblock How fine can fine-tuning be? learning efficient language models.
\newblock In \emph{International Conference on Artificial Intelligence and
  Statistics}, pages 2435--2443. PMLR, 2020.

\bibitem[Sung et~al.(2021)Sung, Nair, and Raffel]{FISHmask}
Yi-Lin Sung, Varun Nair, and Colin~A Raffel.
\newblock Training neural networks with fixed sparse masks.
\newblock In M.~Ranzato, A.~Beygelzimer, Y.~Dauphin, P.S. Liang, and J.~Wortman
  Vaughan, editors, \emph{Advances in Neural Information Processing Systems},
  volume~34, pages 24193--24205. Curran Associates, Inc., 2021.
\newblock URL
  \url{https://proceedings.neurips.cc/paper/2021/file/cb2653f548f8709598e8b5156738cc51-Paper.pdf}.

\bibitem[Zhang et~al.(2021{\natexlab{a}})Zhang, Tay, Zhang, Chan, Luu, Hui, and
  Fu]{zhang2021beyond}
Aston Zhang, Yi~Tay, Shuai Zhang, Alvin Chan, Anh~Tuan Luu, Siu Hui, and Jie
  Fu.
\newblock Beyond fully-connected layers with quaternions: Parameterization of
  hypercomplex multiplications with $1/n $ parameters.
\newblock In \emph{International Conference on Learning Representations},
  2021{\natexlab{a}}.

\bibitem[Karimi~Mahabadi et~al.(2021)Karimi~Mahabadi, Henderson, and
  Ruder]{karimi2021compacter}
Rabeeh Karimi~Mahabadi, James Henderson, and Sebastian Ruder.
\newblock Compacter: Efficient low-rank hypercomplex adapter layers.
\newblock \emph{Advances in Neural Information Processing Systems},
  34:\penalty0 1022--1035, 2021.

\bibitem[Deng et~al.(2022)Deng, Wang, Hsieh, Wang, Guo, Shu, Song, Xing, and
  Hu]{https://doi.org/10.48550/arxiv.2205.12548}
Mingkai Deng, Jianyu Wang, Cheng-Ping Hsieh, Yihan Wang, Han Guo, Tianmin Shu,
  Meng Song, Eric~P. Xing, and Zhiting Hu.
\newblock Rlprompt: Optimizing discrete text prompts with reinforcement
  learning, 2022.
\newblock URL \url{https://arxiv.org/abs/2205.12548}.

\bibitem[Zhong et~al.(2022)Zhong, Gao, Ding, Liu, Zhou, Wang, Yin, and
  Duan]{https://doi.org/10.48550/arxiv.2208.03229}
Wanjun Zhong, Yifan Gao, Ning Ding, Zhiyuan Liu, Ming Zhou, Jiahai Wang, Jian
  Yin, and Nan Duan.
\newblock Improving task generalization via unified schema prompt, 2022.
\newblock URL \url{https://arxiv.org/abs/2208.03229}.

\bibitem[Bari et~al.(2022)Bari, Zhang, Zheng, Shi, Zhu, Joty, and
  Li]{bari2022spt}
M~Saiful Bari, Aston Zhang, Shuai Zheng, Xingjian Shi, Yi~Zhu, Shafiq Joty, and
  Mu~Li.
\newblock Spt: Semi-parametric prompt tuning for multitask prompted learning.
\newblock \emph{arXiv preprint arXiv:2212.10929}, 2022.

\bibitem[Ding et~al.(2022)Ding, Qin, Yang, Wei, Yang, Su, Hu, Chen, Chan, Chen,
  Yi, Zhao, Wang, Liu, Zheng, Chen, Liu, Tang, Li, and
  Sun]{https://doi.org/10.48550/arxiv.2203.06904}
Ning Ding, Yujia Qin, Guang Yang, Fuchao Wei, Zonghan Yang, Yusheng Su,
  Shengding Hu, Yulin Chen, Chi-Min Chan, Weize Chen, Jing Yi, Weilin Zhao,
  Xiaozhi Wang, Zhiyuan Liu, Hai-Tao Zheng, Jianfei Chen, Yang Liu, Jie Tang,
  Juanzi Li, and Maosong Sun.
\newblock Delta tuning: A comprehensive study of parameter efficient methods
  for pre-trained language models, 2022.
\newblock URL \url{https://arxiv.org/abs/2203.06904}.

\bibitem[Jawahar et~al.(2019)Jawahar, Sagot, and
  Seddah]{jawahar-etal-2019-bert}
Ganesh Jawahar, Beno{\^\i}t Sagot, and Djam{\'e} Seddah.
\newblock What does {BERT} learn about the structure of language?
\newblock In \emph{Proceedings of the 57th Annual Meeting of the Association
  for Computational Linguistics}, pages 3651--3657, Florence, Italy, July 2019.
  Association for Computational Linguistics.

\bibitem[Zhang et~al.(2021{\natexlab{b}})Zhang, Tay, Shen, Chan, and
  Zhang]{zhang2021self}
Aston Zhang, Yi~Tay, Yikang Shen, Alvin Chan, and Shuai Zhang.
\newblock Self-instantiated recurrent units with dynamic soft recursion.
\newblock \emph{Advances in Neural Information Processing Systems},
  34:\penalty0 6503--6514, 2021{\natexlab{b}}.

\bibitem[Wang et~al.(2018)Wang, Singh, Michael, Hill, Levy, and
  Bowman]{Wang2018GLUEAM}
Alex Wang, Amanpreet Singh, Julian Michael, Felix Hill, Omer Levy, and
  Samuel~R. Bowman.
\newblock Glue: A multi-task benchmark and analysis platform for natural
  language understanding.
\newblock In \emph{BlackboxNLP@EMNLP}, 2018.

\bibitem[Narayan et~al.(2018)Narayan, Cohen, and
  Lapata]{narayan-etal-2018-dont}
Shashi Narayan, Shay~B. Cohen, and Mirella Lapata.
\newblock Don{'}t give me the details, just the summary! topic-aware
  convolutional neural networks for extreme summarization.
\newblock In \emph{Proceedings of the 2018 Conference on Empirical Methods in
  Natural Language Processing}, pages 1797--1807, Brussels, Belgium,
  October-November 2018. Association for Computational Linguistics.
\newblock \doi{10.18653/v1/D18-1206}.
\newblock URL \url{https://aclanthology.org/D18-1206}.

\bibitem[Bojar et~al.(2016)Bojar, Chatterjee, Federmann, Graham, Haddow, Huck,
  Jimeno~Yepes, Koehn, Logacheva, Monz, Negri, N{\'e}v{\'e}ol, Neves, Popel,
  Post, Rubino, Scarton, Specia, Turchi, Verspoor, and
  Zampieri]{bojar-etal-2016-findings}
Ond{\v{r}}ej Bojar, Rajen Chatterjee, Christian Federmann, Yvette Graham, Barry
  Haddow, Matthias Huck, Antonio Jimeno~Yepes, Philipp Koehn, Varvara
  Logacheva, Christof Monz, Matteo Negri, Aur{\'e}lie N{\'e}v{\'e}ol, Mariana
  Neves, Martin Popel, Matt Post, Raphael Rubino, Carolina Scarton, Lucia
  Specia, Marco Turchi, Karin Verspoor, and Marcos Zampieri.
\newblock Findings of the 2016 conference on machine translation.
\newblock In \emph{Proceedings of the First Conference on Machine Translation:
  Volume 2, Shared Task Papers}, pages 131--198, Berlin, Germany, August 2016.
  Association for Computational Linguistics.
\newblock \doi{10.18653/v1/W16-2301}.
\newblock URL \url{https://aclanthology.org/W16-2301}.

\bibitem[Lin(2004)]{lin-2004-rouge}
Chin-Yew Lin.
\newblock {ROUGE}: A package for automatic evaluation of summaries.
\newblock In \emph{Text Summarization Branches Out}, pages 74--81, Barcelona,
  Spain, July 2004. Association for Computational Linguistics.
\newblock URL \url{https://aclanthology.org/W04-1013}.

\bibitem[Papineni et~al.(2002)Papineni, Roukos, Ward, and
  Zhu]{Papineni2002BleuAM}
Kishore Papineni, Salim Roukos, Todd Ward, and Wei-Jing Zhu.
\newblock Bleu: a method for automatic evaluation of machine translation.
\newblock In \emph{ACL}, 2002.

\end{thebibliography}
\bibliographystyle{unsrtnat}

\newpage

\appendix

\section{More Experimental Results}

\begin{table}[h]
\caption{Average performances (low-compute, low-epoch regime: 100 random models, tuning epochs $=1,2,3,4,20$ for five different blocks) on the GLUE datasets using the T5-base pretrained backbone model. We compare adding different grouping constraints to the $\mathcal{S}_0$ design space.} \label{Tab:Grouping-1}

\small
\begin{center}
\begin{tabular}{c|cccccccc|c}
\toprule
 \textbf{Grouping Patterns}                            & \textbf{SST-2}              & \textbf{MNLI}                & \textbf{QNLI}                & \textbf{QQP}                 & \textbf{RTE}                & \textbf{STS-B}               & \textbf{MRPC}                & \textbf{CoLA}  & \textbf{Avg}               \\ \midrule   \midrule
\multicolumn{10}{c}{1 epochs}
 \\ \midrule
\textbf{Increasing}                           & 73.2           & 63.3           & \textbf{67.8}          & 68.8          & \textbf{63.8}         & 67.2         & 64.1          & 11.0   & \textbf{59.9}      \\ 
Uniform                                      & \textbf{72.8}          & 64.1          & 63.4           & 63.4         & 62.5          & \textbf{69.8}          & 65.8     & 12.1     &59.2      \\ 
Decreasing                              & 72.4           & 63.2        & 65.1          & 69.8          & 59.3           & 62.7         & 63.6         & \textbf{18.7}     &59.4    \\ 
Spindle & 72.6  & \textbf{64.8} & 66.8  & \textbf{71.1} & 62.1  & 62.3 & 64.8 & 12.3  & 59.6 \\ 
Bottleneck         & 72.2          & 63.7           & 65.3         & 68.3          & 61.2          & 63.2         & \textbf{66.6}          & 12.1  &59.0   \\ \midrule \midrule
\multicolumn{10}{c}{2 epochs}
 \\ \midrule
Increasing                           & 76.2           & 69.3           & 73.2         & 76.5          & 65.8          & 72.2          & \textbf{74.0}          & 21.0   &66.0      \\ 
Uniform                                      & 74.8          & 70.9          & \textbf{74.1}          & 75.6          & 66.5          & 73.4         & 71.2           & 22.1     &66.1    \\ 
Decreasing                              & 71.4           & 70.1         & 72.1        & \textbf{76.8}          & 64.3         & 71.7          & 73.6          & 18.7      &64.8  \\ 
\textbf{Spindle} & \textbf{76.6} & \textbf{71.9} & 71.8 & 74.4 & \textbf{67.5} &  \textbf{73.5} & 71.8  & 22.3 & \textbf{66.2}  \\ 
Bottleneck         & 74.2           & 71.1           & 69.6           & 73.3           & 65.2          & 73.3       & 73.6        & \textbf{24.1}  &65.5  \\ \midrule \midrule  
\multicolumn{10}{c}{3 epochs}
 \\ \midrule
Increasing                           & 85.3          & 74.9    & 77.2          & 77.5          & 66.8        & 76.2       & 76.0          & 33.0       &70.8 \\ 
Uniform                                      & 84.8           & 73.7        & 78.1          & 78.6         & 68.5        & 77.8         & 79.2        & 36.1      & 72.1  \\ 
Decreasing                              & 81.9           & 72.1          & 78.3          & 76.7        & 67.3          & 75.9        & 78.6       & 28.7  &69.9       \\ 
\textbf{Spindle} & \textbf{86.9} & \textbf{75.5} & \textbf{79.8} & \textbf{79.4} & \textbf{69.8} & \textbf{78.3} & \textbf{80.1} & \textbf{47.3} &\textbf{74.6} \\ 
Bottleneck         & 84.5          & 74.6          & 76.9         & 78.1           & 69.2          & 76.2           & 78.6         & 32.1 &71.3    \\ \midrule \midrule  
\multicolumn{10}{c}{4 epochs}
 \\ \midrule
Increasing                           & 88.3          & 78.5          & 80.2           & 80.5          & 70.8           & 80.2        & 80.0          & 37.0    &74.4     \\ 
Uniform                                      & 88.8          & 78.9         & 81.9         & 81.5          & 71.5           & 80.8          & 81.4       & 39.1   &75.4     \\ 
Decreasing                              & 87.6           & 74.1         & 80.8          & 81.7          & 79.3          & 78.9           & 79.6          & 38.7     &75.1    \\ 
\textbf{Spindle} & \textbf{89.6} & \textbf{79.8} & \textbf{83.6} & \textbf{82.8}  &\textbf{71.8} & \textbf{81.3} & \textbf{82.1} & \textbf{39.3} &\textbf{76.3} \\ 
Bottleneck         & 86.5           & 77.6           & 82.7           & 81.1           & 70.2           & 70.9          & 81.6           & 36.1  &73.3    \\  \midrule \midrule

\multicolumn{10}{c}{20 epochs}
 \\ \midrule
Increasing                            & 92.3     & 83.3        & 86.2          & 82.5        & 71.8        & 82.2     & 84.0       & 51.0     &79.1   \\ 
Uniform                                     & 92.8       & 83.9           & 86.1          & 83.6       & 72.5     & 83.8      & 84.2         & 52.1 &79.9   \\ 
Decreasing                               & 91.4         & 82.1         & 85.1          & 83.1        & 69.3        & 81.7       & 83.6         & 48.7    &78.1      \\ 

\textbf{Spindle} & \textbf{93.6} & \textbf{84.8} & \textbf{87.8} & \textbf{84.4} & \textbf{73.5} & \textbf{84.3} & \textbf{85.8} & \textbf{52.3} &\textbf{80.8} \\ 
Bottleneck          & 92.1        & 82.6        & 85.6           & 83.3        & 71.2         & 83.2        & 84.6        & 52.1    &79.3  \\ \bottomrule   

\end{tabular} 
\end{center}
\end{table}

\begin{table}[h]
\caption{Average performances (low-compute, low-epoch regime: 100 random models, 3 tuning epochs) on the GLUE datasets using the T5-base pretrained backbone model. We compare adding different $G_1$ strategy assignment constraints to the $\mathcal{S}_3$ design space.} \label{Tab:$G_1$}
\begin{center}
\small
\begin{tabular}{c|cccccccc|c} \toprule
\textbf{Strategy Assignment}     & \multicolumn{1}{l}{\textbf{SST-2}} & \multicolumn{1}{l}{\textbf{MNLI}} & \multicolumn{1}{l}{\textbf{QNLI}} & \multicolumn{1}{l}{\textbf{QQP}} & \multicolumn{1}{l}{\textbf{RTE}} & \multicolumn{1}{l}{\textbf{STS-B}} & \multicolumn{1}{l}{\textbf{MRPC}} & \multicolumn{1}{l}{\textbf{CoLA}} &\multicolumn{1}{|l}{\textbf{Avg}} \\ \midrule \midrule
$G_1$-Adapter (A)                       & 89.8                               & 83.5                              & 84.9                              & 80.8                             & 72.5                             & 80.8                               & \textbf{78.5}                     & \textbf{37.7}          &76.1           \\
$G_1$-Prefix (P)                         & 89.3                               & 83.1                              & 84.4                              & 80.1                             & 70.1                             & 80.0                                 & 77.6                              & 33.0                            &74.7    \\
$G_1$-BitFit (B)                         & 89.0                                 & 82.9                              & 84.1                              & 81.4                             & 72.0                               & 81.1                               & 77.0                                & 30.8            &74.8                  \\
$G_1$-LoRA (L)                           & 89.9                               & 83.6                              & 85.0                                & 81.1                             & 71.8                             & 81.0                                 & 78.8                              & 35.3                      &75.8        \\
$G_1$-(P, L)                  & 89.1                               & 82.8                              & 85.1                              & 81.2                             & 71.9                             & 81.5                               & 79.1                                & 35.0    &  75.7                             \\
$G_1$-(A, P)               & 89.8                               & 82.8                              & 84.8                              & 81.1                             & 72.2                             & 81.3                               & 79.2                              & 36.4    & 75.9                         \\
\textbf{$G_1$-(A, L)}        & 89.6                     & \textbf{83.8}                     & \textbf{85.6}                     & 81.3                    & \textbf{72.9}                    & \textbf{81.7}                      & \textbf{79.5}                     & \textbf{36.8}      &\textbf{76.4}               \\
$G_1$-(A, P, L)         & 89.6                               & 83.5                              & 85.2                              & 81.5                             & 72.2                             & 81.4                               & 79.2                              & 35.2   &75.9                           \\
$G_1$-(P, B, L)          & 89.3                               & 83.6                              & 85.5                              & 81.6                             & 72.3                             & 81.0                                 & 78.8                              & 35.7    &76.0                          \\
$G_1$-(A, P, B)       & 89.2                               & 83.3                              & 84.8                      & \textbf{81.8}                    & 72.5                    & 81.1                      & 78.6                              & 35.6             & 75.8                \\
$G_1$-(A, B, L)         & 89.8                               & 83.4                              & 84.8                              & 81.1                             & 72.6                             & 81.6                               & 79.4                              & 34.8    &75.9                          \\
$G_1$-(A, P, B, L) & \textbf{90.0}                        & 83.1                     & 85.3                     & 81.6                             & 72.6                             & 81.4                               & 79.2                              & 36.5   &76.1  \\ \bottomrule                           
\end{tabular}
\end{center}
\end{table}

\begin{table}[h]
\caption{Average performances (low-compute, low-epoch regime: 100 random models, 3 tuning epochs) on the GLUE datasets using the T5-base pretrained backbone model. We compare adding different $G_2$ strategy assignment constraints with $G_1$-(L, A) to the $\mathcal{S}_3$ design space.
} \label{Tab:$G_2$}
\begin{center}

\small
\begin{tabular}{c|cccccccc|c} \toprule
\textbf{Strategy Assignment}     & \multicolumn{1}{l}{\textbf{SST-2}} & \multicolumn{1}{l}{\textbf{MNLI}} & \multicolumn{1}{l}{\textbf{QNLI}} & \multicolumn{1}{l}{\textbf{QQP}} & \multicolumn{1}{l}{\textbf{RTE}} & \multicolumn{1}{l}{\textbf{STS-B}} & \multicolumn{1}{l}{\textbf{MRPC}} & \multicolumn{1}{l}{\textbf{CoLA}} & \multicolumn{1}{|l}{\textbf{Avg}} \\ \midrule \midrule
$G_2$-Adapter (A)                         & 91.6                               & 84.3                              & 85.5                              & \textbf{82.3}                            & 73.5                             & 82.8                               & 81.3                     & 38.8                & 77.5    \\
$G_2$-Prefix (P)                          & 89.6                               & 84.0                              & 86.5                              & 81.5                            & 73.3                             & 82.5                                 & 80.5                              & 36.2                             &76.7  \\
$G_2$-BitFit (B)                         &91.2          & 83.6          & 85.7          & 82.9          & 72.6          & 82.6          & 80.8          & 33.1     &76.5     \\
$G_2$-LoRA (L)                            & 91.4          & 84.4          & 86.1          & 82.0            & 72.8          & 81.8          & 81.6          & 39.8    &77.4      \\
$G_2$-(P, L)                  & 91.6          & 84.6          & 86.8          & 81.8          & 73.8          & 82.8          & 82.0            & 38.5   &77.7       \\
\textbf{$G_2$-(A, P)}               & \textbf{92.2} & \textbf{84.2} & \textbf{87.1} & 82.2          & \textbf{74.4} & 83.0            & \textbf{82.5} & 40.8      &\textbf{78.3}    \\
$G_2$-(A, L)        &92.0            & 84.4          & 86.5          & 81.8          & 73.6          & 82.6          & 82.2          & 40.1     &77.9     \\
$G_2$-(A, P, L)         & 91.8          & 84.8          & 86.8          & 81.8          & 74.1          & 83.0            & 82.1          & 37.9     &77.7     \\
$G_2$-(P, B, L)          & 91.6          & 84.1          & 87.1          & 82.0            & 74.0           & 82.9          & 82.4          & 35.8       &77.4   \\
$G_2$-(A, P, B)       & 91.8          & 84.2          & 86.8          & 82.1          & 73.7          & \textbf{83.3} & 82.2          & 41.2   &78.1       \\
$G_2$-(A, B, L)         & \textbf{92.2} & 84.3          & 86.1          & 82.0            & 74.1          & 83.2          & 82.0            & 37.6      &77.6    \\
$G_2$-(A, P, B, L) & 92.0            & 84.1          & 87.0            & 81.9          & 74.2          & 83.1          & 81.3          & \textbf{42.4} &78.1  \\ \bottomrule                           
\end{tabular} 
    
\end{center}
\end{table}

\begin{table}[h]
\caption{Average performances (low-compute, low-epoch regime: 100 random models, 3 tuning epochs) on the GLUE datasets using the T5-base pretrained backbone model. We compare adding different $G_3$ strategy assignment constraints  with $G_1$-(L, A)  -- $G_2$-(P, A)  to the $\mathcal{S}_3$ design space.
} \label{Tab:$G_3$}

\begin{center}

\small
\begin{tabular}{c|cccccccc|c} \toprule
\textbf{Strategy Assignment}     & \multicolumn{1}{l}{\textbf{SST-2}} & \multicolumn{1}{l}{\textbf{MNLI}} & \multicolumn{1}{l}{\textbf{QNLI}} & \multicolumn{1}{l}{\textbf{QQP}} & \multicolumn{1}{l}{\textbf{RTE}} & \multicolumn{1}{l}{\textbf{STS-B}} & \multicolumn{1}{l}{\textbf{MRPC}} & \multicolumn{1}{l}{\textbf{CoLA}} &\multicolumn{1}{|l}{\textbf{Avg}}  \\ \midrule \midrule
$G_3$-Adapter (A)                         & 92.5          & 85.3          & 87.5          & \textbf{83.3} & 73.9          & 84.0            & 83.8          & \textbf{44.9}  &79.4 \\
$G_3$-Prefix (P)                         &91.5          & 84.7          & 86.7          & 82.6          & 74.2          & 83.8          & 82.9          & 40.5        &78.4  \\
$G_3$-BitFit (B)                         &91.9          & 84.3          & 87.0            & 82.0            & 73.6          & 84.1          & 83.3          & 36.1       &77.8   \\
$G_3$-LoRA (L)                            & 92.8          & 85.4          & 87.8          & 83.5          & 74.7          & 82.4          & 84.0            & 44.0         &79.3   \\
$G_3$-(P, L)                  & 93.0            & 85.2          & 88.3          & 83.8          & 75.2          & 84.4          & 84.2          & 37.9     & 79.0    \\
$G_3$-(A, P)               & 92.4          & 85.6          & 88.1            & 83.6        & 75.0            & 84.2          & 84.0            & 41.8       & 79.3  \\
$G_3$-(A, L)        &92.0            & 85.9          & 88.2          & 83.1          & 75.3          & 84.3          & 83.9          & 42.2      &79.4    \\
$G_3$-(A, P, L)         & 92.6          & 86.0            & 87.5          & 83.4        & 75.6          & 84.6          & 83.5          & 43.9    &79.6      \\
$G_3$-(P, B, L)         & 92.7          & 85.8          & 87.2          & 83.7        & 75.2          & 84.5          & 83.8          & 40.8      & 79.2    \\
\textbf{$G_3$-(A, P, B)}       & 93.3          & \textbf{85.8} & \textbf{88.6} & \textbf{84.0} & 75.5          & \textbf{84.9} & 84.1          & 42.1     & \textbf{79.8}    \\
$G_3$-(A, B, L)         & \textbf{93.7} & 86.5          & 88.0            & 83.2        & \textbf{75.8} & 84.2          & 84.2          & 39.7     &79.4     \\
$G_3$-(A, P, B, L) & 93.3          & 85.6          & 87.7          & 83.8        & 75.2          & 84.3          & \textbf{84.4} & 41.6      &  79.4   \\ \bottomrule                           
\end{tabular} 
\end{center}
\end{table}

\begin{table}[h]
\caption{Average performances (low-compute, low-epoch regime: 100 random models, 3 tuning epochs) on the GLUE datasets using the T5-base pretrained backbone model. We compare adding different $G_4$ strategy assignment constraints  with $G_1$-(A, L)  -- $G_2$-(A, P) -- $G_3$-(A, P, B)  to the $\mathcal{S}_3$ design space.
} \label{Tab:$G_4$}
\begin{center}
    
\small
\begin{tabular}{c|cccccccc|c} \toprule
\textbf{Strategy Assignment}     & \multicolumn{1}{l}{\textbf{SST-2}} & \multicolumn{1}{l}{\textbf{MNLI}} & \multicolumn{1}{l}{\textbf{QNLI}} & \multicolumn{1}{l}{\textbf{QQP}} & \multicolumn{1}{l}{\textbf{RTE}} & \multicolumn{1}{l}{\textbf{STS-B}} & \multicolumn{1}{l}{\textbf{MRPC}} & \multicolumn{1}{l}{\textbf{CoLA}} &\multicolumn{1}{|l}{\textbf{Avg}} \\ \midrule \midrule
$G_4$-Adapter (A)                         & 93.8          & 85.8          & 88.6          & 84.8          & 76.3          & 85.8          & 86.0            & \textbf{48.5}  &81.2 \\
$G_4$-Prefix (P)                         &93.5          & 85.2          & 88.3          & 83.6          & 76.8          & 85.3          & 85.6          & 44.8     &80.3     \\
$G_4$-BitFit (B)                       &94.1          & 85.3          & 88.9          & 84.4          & 77.1          & 85.4          & 86.2          & 46.1        &80.9  \\
$G_4$-LoRA (L)                               & 94.0            & 86.0            & 89.2          & 85.0            & 77.2          & 85.5          & 85.8          & 47.7   &81.3       \\
$G_4$-(P, L)                  & 94.3          & 86.2          & 89.3          & 85.8          & 78.0            & 86.0            & 88.2          & 47.2    &81.8      \\
$G_4$-(A, P)               & 94.1            & 86.2            & 89.6          & 85.4          & 77.9          & 86.2          & 86.9          & 45.3      & 81.4   \\
$G_4$-(A, L)        &94.2          & 85.9          & 89.2          & 85.5          & 77.8          & 86.2          & 88.0            & 46.8     &81.7     \\
$G_4$-(A, P, L)       &94.1          & 85.8          & 88.8          & 85.7          & 77.4          & 86.5          & 87.9          & 44.8     &81.3     \\
\textbf{$G_4$-(P, B, L)}          &\textbf{94.6} & \textbf{86.4} & \textbf{90.4} & \textbf{86.1} & 78.2          & \textbf{86.8} & \textbf{88.5} & 47.2    &\textbf{82.3}      \\
$G_4$-(A, P, B)       &94.5          & 86.0            & 89.6          & 86.0            & 78.0            & 86.2          & 88.1          & 44.8       &81.6   \\
$G_4$-(A, B, L)         & 94.3          & \textbf{86.4} & 89.2          & 85.6          & 78.2          & 86.4          & 88.3          & 46.6       &81.9   \\
$G_4$-(A, P, B, L) &94.2          & 86.2          & 89.2          & 85.9          & \textbf{78.5} & 86.1          & 88.0            & 45.3     &81.6      \\ \bottomrule                          
\end{tabular} 
\end{center}
\end{table}

\begin{table}[h]
\caption{Average performances (low-compute, low-epoch regime: 100 random models, 3 tuning epochs) on the GLUE datasets using the T5-3b pretrained backbone model. We compare adding different layer grouping constraints to the $\mathcal{S}_0$ design space.} \label{Tab:Grouping-3-3b}
\begin{center}

\small
\begin{tabular}{c|cccccccc|c}
\toprule
\textbf{Grouping Patterns}                             & \textbf{SST-2}              & \textbf{MNLI}                & \textbf{QNLI}                & \textbf{QQP}                 & \textbf{RTE}                & \textbf{STS-B}               & \textbf{MRPC}                & \textbf{CoLA}       & \textbf{Avg}          \\ \midrule  \midrule 

$\mathcal{S}_0$-models & 80.3 & 72.1 & 74.7  &72.8  & 76.9 & 75.2 & 71.0 & 32.2 & 69.4 \\

\midrule  \midrule 

Increasing                           & 84.4     & 75.7 & 83.0          & 78.3        & 82.7       & 80.3    & 76.3       & 42.1    &75.3    \\ 
Uniform                                      & 86.8       & 77.1           & 82.6          & 76.2       & 83.8     & \textbf{81.6}      & 77.3         &\textbf{48.9}  &76.8  \\ 
Decreasing                              & 83.2         & 74.3         & 81.8          & 77.3        & 82.8        & 79.9       & 76.5         & 40.8       & 74.5  \\ 
\textbf{Spindle} & \textbf{88.6} & \textbf{78.8} & \textbf{83.7} &  77.7 & \textbf{84.2} & 80.9 & \textbf{78.3} & 44.6 &\textbf{77.1} \\ 
Bottleneck         & 86.3        &77.0      & 82.2           & 75.6        & 83.3        &80.2        & 77.1       & 41.5  &  75.4  \\ \bottomrule       
\end{tabular} 
\end{center}
\end{table}

\begin{table}[h]
\caption{Average performances (low-compute, low-epoch regime: 100 random models, 3 tuning epochs) on the GLUE datasets using the T5-3b pretrained backbone model. We compare adding different layer parameter constraints to the $\mathcal{S}_1$ design space.} \label{Tab:Allocate-3b}
\begin{center}
\small
\begin{tabular}{c|cccccccc|c}
\toprule
\textbf{Parameter Allocation} & \textbf{SST-2}                          & \textbf{MNLI}                           & \textbf{QNLI}       & \textbf{QQP}        & \textbf{RTE}        & \textbf{STS-B}      & \textbf{MRPC}       & \textbf{CoLA}    & \textbf{Avg}      \\ \midrule \midrule
Increasing   & 90.3                           & 79.3                     & \textbf{84.9} & 79.3       & 85.2   & \textbf{82.8}         & \textbf{79.2} & 50.1  &78.9 \\
\textbf{Uniform}        & \textbf{90.6}   & \textbf{80.8} & 84.6 & \textbf{79.7} & \textbf{85.5} & 82.4  & 78.9  &\textbf{50.8}   &\textbf{79.1}   \\
Decreasing        & 88.6                           & 78.2                             & 83.5         & 78.1           & 84.4           & 81.5         & 78.1       & 49.6 &77.7 \\ \bottomrule
\end{tabular} 
\end{center}
\end{table}

\begin{table}[h]
 \caption{Average performances (low-compute, low-epoch regime: 100 random models, 3 tuning epochs) on the GLUE datasets using the T5-3b pretrained backbone model. We compare adding different tuning groups constraints to the $\mathcal{S}_2$ design space. \label{Tab:Tunable-3b} 
}
\begin{center}
\small
\begin{tabular}{c|cccccccc|c}
\toprule
\textbf{Tunable Groups} & \textbf{SST-2}      & \textbf{MNLI}       & \textbf{QNLI}        & \textbf{QQP}        & \textbf{RTE}         & \textbf{STS-B}      & \textbf{MRPC}        & \textbf{CoLA} &\textbf{Avg}       \\ \midrule \midrule
$G_1$                      & 88.3                               & 78.3                              & 82.2                              & 77.4                             & 82.1                             & 80.7                               & 76.1                              & 49.4    &76.8       \\
$G_2$                     & 89.1                               & 78.8                              & 82.1                              & 77.2                             & 82.3                             & 81.2                               & 76.4                     & 49.6             &77.1   \\
$G_3$                     & 89.6                               & 78.5                              & 82.6                              & 78.1                             & 83.8                             & 81.9                               & 77.4                              & 48.7             &77.5      \\
$G_4$                       & 89.8                               & 79.3                              & 82.7                              & 77.9                             & 83.5                             & 81.9                               & 77.9                              & 48.5   &77.1          \\
$G_1$, $G_2$                 & 90.1                               & 80.2                              & 83.4                              & 78.5                             & 84.3                             & 82.4                               & 78.5                              & 51.1    &78.5                                 \\
$G_3$, $G_4$                  &  90.5                               & 80.6                              & 83.8                              & 78.7                             & 84.2                             & 83                                 & 78.2                              & 50.3  &78.6           \\
$G_1$, $G_2$, $G_3$             & 90.6                               & 80.3                              & 84.9                              & 79.3                             & 84.7                             & 82.9                               & 79.3                              & 50.2  &79.0           \\
$G_2$, $G_3$, $G_4$             & 90.8                               & 80.9                              & 84.6                              & 79.1                             & 85.1                             & 83.1                               & 79.1                              & 49.2 &78.9            \\
$\boldsymbol{G_1}$, $\boldsymbol{G_2}$, $\boldsymbol{G_3}$, $\boldsymbol{G_4}$ & \textbf{91.1} & \textbf{81.4} & \textbf{85.2} & \textbf{80.4} & \textbf{85.9} & \textbf{83.5} & \textbf{80.0} & \textbf{51.6} & \textbf{79.9}        \\ \bottomrule
\end{tabular}
\end{center}
\end{table}

\begin{table}[h]
\caption{Average performances (low-compute, low-epoch regime: 100 random models, 3 tuning epochs) on the GLUE datasets using the T5-3b pretrained backbone model. We compare adding different  strategy assignment constraints  following the process in Section \ref{subsec:s4}. 
} \label{Tab:g-3b}
\begin{center}
\centering
\small
\begin{tabular}{c|cccccccc|c} \toprule
\textbf{Strategy Assignment}     & \multicolumn{1}{l}{\textbf{SST-2}} & \multicolumn{1}{l}{\textbf{MNLI}} & \multicolumn{1}{l}{\textbf{QNLI}} & \multicolumn{1}{l}{\textbf{QQP}} & \multicolumn{1}{l}{\textbf{RTE}} & \multicolumn{1}{l}{\textbf{STS-B}} & \multicolumn{1}{l}{\textbf{MRPC}} & \multicolumn{1}{l}{\textbf{CoLA}} &\multicolumn{1}{|l}{\textbf{Avg}}\\ \midrule \midrule
$G_1$-Adapter (A)                        & 91.1                               & 81.4                              & 86.1                              & 80.5                             & 86.7                             & 83.3                               & 80.1                              & 50.8                    &80.0          \\
$G_1$-Prefix (P)                         & 90.8                               & 81.1                              & 85.5                              & 80.2                             & 86.2                             & 83.1                               & 79.8                              & 50.2                 & 79.6            \\
$G_1$-BitFit (B)                         & 90.2                               & 81.3                              & 85.1                              & 79.6                             & 85.8                             & 82.8                               & 79.6                              & 49.5                           &79.2   \\
$G_1$-LoRA  (L)                          & 91.4                               & 81.9                              & 86.2                              & 80.8                             & 86.4                             & 83.9                               & 80.8                              & 49.6                          &80.0    \\
\textbf{$G_1$-(P, L)}         & \textbf{91.8}                      & \textbf{82.9}                     & \textbf{86.8}                     & 81.3                             & \textbf{87.1}                    & 84.2                               & \textbf{81.6}                     & 52.3                           &\textbf{81.0}   \\
$G_1$-(A, P)               & 91.3                               & 81.9                              & 86.4                              & 81.1                             & 85.6                             & 83.7                               & 80.7                              & \textbf{52.8}                   &80.1   \\
$G_1$-(A, L)       & 91.6                               & 82.3                              & 86.1                              & \textbf{81.5}                    & 85.8                             & \textbf{84.9}                      & 81.5                              & 51.8                           &80.6   \\
$G_1$-(A, P, L)         & 91.1                               & 81.7                              & 85.8                              & 81.2                             & 86.4                             & 84.2                               & 80.9                              & 52.3                            &80.4  \\
$G_1$-(P, B, L)          & 91.5                               & 82.8                              & 86.3                              & 81.4                             & 86.1                             & 83.6                               & 81.2                              & 51.5                             &80.5 \\
$G_1$-(A, P, B)       & 91.3                               & 82.3                              & 86.7                              & 80.8                             & 86.8                             & 84.3                               & 80.7                              & 51.8                            &80.5  \\
$G_1$-(A, B, L)         & 91.7                               & 82.5                              & 86.2                              & 81.3                             & 86.3                             & 84.6                               & 81.3                              & 51.7                            & 80.7  \\
$G_1$-(A, P, B, L) & 91.6                               & 82.3                              & 86.2                              & 81.1                             & 86.6                             & 84.2                               & 81.1                              & 51.1     &80.5                        
  \\  \midrule  \midrule
$G_2$-Adapter (A)                         & 92.1                               & 82.5                              & 86.4                              & 81.8                             & 87.2                             & 84.8                               & 81.8                              & 53.8                           &81.3   \\
$G_2$-Prefix (P)                          & 91.8                               & 83.1                              & 87.2                              & 81.6                             & 86.2                             & 84.4                               & 81.1                              & 52.8                          &81.0    \\
$G_2$-BitFit  (B)                        & 91.2                               & 82.1                              & 86.4                              & 81.1                             & 86.3                             & 84.6                               & 80.3                              & 53.1                           &80.6   \\
$G_2$-LoRA  (L)                          & 92.6                               & 82.9                              & 87.5                              & 81.3                             & 87.4                             & 85.1                               & 81.9                              & 52.2                           &81.4   \\
$G_2$-(P, L)                  & 91.6                               & 82.7                              & 87.6                              & 81.6                             & \textbf{87.8}                    & 85.3                               & 82.1                              & 52.8                          &81.4    \\
$G_2$-(A, P)               & 92.1                               & 83.3                              & 87.5                              & 81.9                             & 87.4                             & 85.5                               & 81.8                              & 53.1                          &81.5    \\
\textbf{$G_2$-(A, L)}        & 92.5                               & \textbf{83.7}                     & \textbf{88.1}                     & \textbf{82.2}                    & 87.4                             & \textbf{85.7}                      & \textbf{82.9}                     & 53.6                            &\textbf{82.1}  \\
$G_2$-(A, P, L)         & 92.3                               & 83.4                              & 87.4                              & 81.6                             & 87.1                             & 85.3                               & 81.4                              & 53.2                         &81.4     \\
$G_2$-(P, B, L)          & 91.8                               & 83.1                              & 87.4                              & 81.5                             & 87.2                             & 85.1                               & 82.7                              & 53.8                           &81.5   \\
$G_2$-(A, P, B)       & 91.5                               & 82.6                              & 87.8                              & 81.3                             & 86.5                             & 85.2                               & 82.1                              & \textbf{54.2}                &81.4     \\
$G_2$-(A, B, L)         & 92.6                               & 83.5                              & 87.2                              & 82                               & 87.3                             & 86.5                               & 82.5                              & 52.8                            &81.8  \\
$G_2$-(A, P, B, L) & \textbf{92.8}                      & 83.2                              & 87.6                              & 81.6                             & 87.5                             & 85.5                               & 82.4                              & 51.2  &81.5 \\    \midrule  \midrule

$G_3$-Adapter (A)                        & 92.6                               & 84.1                              & 88.3                              & 81.8                             & 87.8                             & 85.4                               & 82.8                              & 55.2                           &82.2   \\
$G_3$-Prefix (P)                         & 92.1                               & 83.3                              & 87.6                              & 81.4                             & 87.1                             & 85.4                               & 82.6                              & 53.5                           &81.6   \\
$G_3$-BitFit  (B)                        & 92.4                               & 83.9                              & 88.4                              & 82.1                             & 87.2                             & 85.8                               & 82.4                              & 53.3                            &81.9  \\
$G_3$-LoRA  (L)                          & 93.1                               & 84.3                              & 87.7                              & 82.4                             & 87.8                             & 86.2                               & 83.1                              & 54.3                          &82.3    \\
$G_3$-(P, L)                  & 92.8                               & 84.1                              & 88.7                              & 82.6                             & 88.2                             & 86.2                               & 83.3                              & 54.7                        &82.6      \\
$G_3$-(A, P)               & 93.1                               & 83.8                              & 89.1                              & 82.3                             & 88.1                             & 85.8                               & 82.6                              & 55.1                          &82.5    \\
$G_3$-(A, L)                 & 92.7                               & 84.5                              & 88.4                              & 82.8                             & 88.2                             & 86.1                               & 83.5                              & 54.6                           &82.6   \\
$G_3$-(A, P, L)         & 92.8                               & 84.6                              & 88.1                              & 82.5                             & 87.7                             & 85.5                               & 83.2                              & 53.8                            &82.3  \\
\textbf{$G_3$-(P, B, L)} & \textbf{93.6}                      & \textbf{84.9}                     & \textbf{89.3}                     & \textbf{83.1}                    & 88.2                             & \textbf{86.5}                      & 83.9                              & \textbf{55.8}                 &\textbf{83.2}    \\
$G_3$-(A, P, B)       & 93.3                               & 83.9                              & 88.5                              & 82.2                             & 88.4                             & 86.2                               & 83.5                              & 55.3                           &82.6   \\
$G_3$-(A, B, L)         & 93.4                               & 84.2                              & 88.9                              & 82.6                             & 87.8                             & 85.8                               & \textbf{84.2}                     & 54.9                          &82.7    \\
$G_3$-(A, P, B, L) & 92.2                               & 84.4                              & 88.7                              & 82.3                             & \textbf{88.5}                    & 86.2                               & \textbf{84.2}                     & 54.2                      &82.5  \\    \midrule  \midrule

$G_4$-Adapter (A)                           & 92.8                               & 85.2                              & 89.1                              & 83.5                             & 87.8                             & 86.5                               & 84.2                              & 56.3                           &83.2   \\
$G_4$-Prefix (P)                             & 92.8                               & 84.6                              & 89.5                              & 82.6                             & 87.4                             & 86.5                               & 83.8                              & 55.8                           &82.8   \\
$G_4$-BitFit (B)                            & 93.8                               & 84.9                              & 89.5                              & 83.3                             & 88.7                             & 86.8                               & 84.4                              & 55.2                       &83.3       \\
$G_4$-LoRA (L)                               & 93.3                               & 84.7                              & 89.3                              & 82.7                             & 88.3                             & 86.2                               & 82.7                              & 54.7                         &82.7     \\
$G_4$-(P, L)                     & 93.8                               & 85.3                              & 89.6                              & 83.6                             & 88.6                             & 86.8                               & 84.6                              & 56.3                         & 83.5    \\
$G_4$-(A, P)                  & 93.8                               & 84.9                              & 89.8                              & 84.3                             & 88.5                             & 86.6                               & 84.8                              & 56.7                         & 83.6    \\
$G_4$-(A, L)                    & 93.7                               & 85.6                              & 89.5                              & 84.1                             & 88.2                             & 86.6                               & 85.2                              & 55.4                         &83.5     \\
$G_4$-(A, P, L)            & 94.2                               & 85.2                              & 89.6                              & 83.9                             & 88.2                             & 86.4                               & 84.9                              & 55.9                            &83.5   \\
$G_4$-(P, B, L)             & 93.8                               & \textbf{85.9}                     & 89.8                              & 83.6                             & 88.6                             & 86.9                               & 85.2                              & 56.3                          &83.7    \\
\textbf{$G_4$-(A, P, B)} & \textbf{94.4}                      & 85.7                              & \textbf{90.1}                     & \textbf{84.8}                    & \textbf{88.9}                    & \textbf{87.2}                      & 85.3                              & \textbf{57.3}                  &\textbf{84.2}   \\
$G_4$-(A, B, L)            & 93.8                               & 85.3                              & 89.5                              & 84.1                             & 88.8                             & 86.7                               & \textbf{85.5}                     & 56.6                           & 83.7  \\
$G_4$-(A, P, B, L)    & 94.1                               & 85.4                              & 89.7                              & 84.4                             & 88.5                             & 86.5                               & 85.2                              & 56.8  &83.8    \\    
\bottomrule                           
\end{tabular}
\end{center}
\end{table}

\begin{table}[t]
\caption{Average performances (low-compute, low-epoch regime: 100 random models, 3 tuning epochs) on the GLUE datasets using the T5-base pretrained backbone model. We compare adding different layer grouping constraints to the $\mathcal{S}_0$ design space. Layer grouping is based on 8 groups.} \label{Tab:Grouping-8-groups}
\begin{center}
\small
\begin{tabular}{c|cccccccc|c}
\toprule
\textbf{Layer Grouping}                             & \textbf{SST-2}              & \textbf{MNLI}                & \textbf{QNLI}                & \textbf{QQP}                 & \textbf{RTE}                & \textbf{STS-B}               & \textbf{MRPC}                & \textbf{CoLA } &\textbf{Avg}              \\ \midrule  \midrule 

 $\mathcal{S}_0$-models &76.9 & 70.1 & 72.5 & 73.3 & 63.6 & 71.7 & 73.8  & 24.3 & 65.7 \\

 \midrule 
Increasing                           & $83.2$          & $74.1$    & $76.6$          & $77.1$          & $67.7$        & 76.8       & 74.7          & 30.0        & 70.0 \\ 
Uniform                                      & 83.6           & 73.4        & 78.0          & 77.9         & 68.2        & 76.4         & 78.6        & 34.2         & 71.3 \\ 
Decreasing                              & 80.3           & 71.6          & 77.4          & 75.5        & 67.0          & 75.3        & 77.2       & 26.4        & 68.9 \\ 
\textbf{Spindle} & \textbf{ 86.2 } & \textbf{ 74.3 } & \textbf{ 79.1 } & \textbf{ 78.6 } & \textbf{ 68.5 } & \textbf{ 77.4 } & \textbf{ 79.5} & \textbf{ 35.1 } & \textbf{72.3} \\ 
Bottleneck         & 83.2          & 73.1          & 75.8         & 77.6           & 67.9          & 75.3           & 78.2         & 31.4    & 70.3 \\ \bottomrule   
\end{tabular} 
\end{center}

\end{table}

\section{General Effectiveness on SuperGLUE with XLNet Backbones}
\label{subsec:effective-xlnet}
We also directly use the 
$\mathcal{S}_4$-model and $\mathcal{S}_4$-3b-model (adopting design patterns discovered using T5-base and T5-3b) to fine-tune the XLNet-base and XLNet-large pretrained backbone models without any extra discovery process.
We keep all the other settings the same and evaluate them on SuperGLUE datasets. 
Table~\ref{Tab:superglue} reiterates the fact that our PEFT design patterns  discovered from T5 models are generelizable to the XLNet backbone models
and outperform the investigated methods in other tasks (SuperGLUE) with no additional discovery process.

\begin{table}[t]
 \caption{Performances of different tuning methods on the SuperGLUE datasets using the XLNet-base (upper part) and XLNet-large (lower part) pretrained backbone models, respectively. The results are averaged over 10 random runs. The $\mathcal{S}_4$-model and $\mathcal{S}_4$-3b-model perform significantly better than the second-best PEFT methods in all the eight datasets at the significance level $p<0.05$ (*) or even $p<0.01$ (**). } \label{Tab:superglue}
 \begin{center}
\scriptsize
\begin{tabular}{c|cccccccc|c} \toprule
\textbf{Method}     & \multicolumn{1}{c}{\textbf{BoolQ}} & \multicolumn{1}{c}{\textbf{CB}} & \multicolumn{1}{c}{\textbf{COPA}} & \multicolumn{1}{c}{\textbf{MultiRC}} & \multicolumn{1}{c}{\textbf{ReCoRD}} & \multicolumn{1}{c}{\textbf{RTE}} & \multicolumn{1}{c}{\textbf{WiC}} & \multicolumn{1}{c}{\textbf{WSC}} & \multicolumn{1}{|c}{\textbf{Average}} \\ \midrule \midrule
Adapter           & 72.8          & 71.3/78.0         & 64.0          & 67.0/24.5          & 71.0/71.8          & 76.2          & 65.0          & 60.8          & 66.2          \\
Prefix            & 72.0            & 70.5/77.0          & 63.3          & 66.4/23.8          & 69.9/71.0          & 75.5          & 64.4          & 60.8            & 65.9          \\
BitFit            & 71.8          & 70.0/76.2          & 62.8          & 65.8/22.6          & 69.4/70.6          & 74.5          & 64.8          & 60.6          & 65.2          \\
\underline{LoRA}              & 72.2          & 71.1/77.8          & 64.7          & 67.4/24.8          & 70.8/71.3          & 76.8          & 65.1          & 61.1          & 66.4          \\ 
\textbf{$\mathcal{S}_4$-model} & \textbf{$\mathbf{73.8}^{**}$} & \textbf{$\mathbf{71.7/78.4}^{*}$} & \textbf{$\mathbf{65.9}^{**}$}          & \textbf{$\mathbf{68.2/25.5}^{**}$}          & \textbf{$\mathbf{71.1/72.0}^{*}$} & \textbf{$\mathbf{78.4}^{**}$}          & \textbf{$\mathbf{65.8}^{*}$} & \textbf{$\mathbf{62.6}^{*}$} & \textbf{67.5} 

\\ \midrule \midrule

Adapter            & 74.4                           & 71.4/81.1                              & 67.4                              & 68.8/26.4                             & 71.7/72.4                             & 80.8                                 & 68.0                             & 64.6                              & 68.8                     \\
Prefix           & 72.4                               & 70.0/78.3                              & 66.9                              & 68.8/25.8                             & 70.9/71.2                             & 78.8                               & 66.9                              & 64.0                              & 67.7                   \\
BitFit              & 71.1                               & 70.7/79.8                              & 68.0                              & 68.6/25.4                             & 71.1/71.6                             & 80.4                               & 67.2                             & 64.3                              & 68.1                 \\
\underline{LoRA}             & 74.1                               & 72.1/80.9                              & 67.9                              & 69.1/26.8                             & 72.0/72.8                             & 81.0                               & 67.8                              & 64.4                              & 69.0                  \\
\textbf{$\mathcal{S}_4$-3b-model}  & \textbf{$\mathbf{76.8}^{**}$}                               & \textbf{$\mathbf{74.6/81.9}^{**}$}                     & \textbf{$\mathbf{68.6}^{**}$}                     & \textbf{$\mathbf{69.5/27.1}^{*}$}                             & \textbf{$\mathbf{72.4/73.3}^{*}$}                    & \textbf{$\mathbf{81.2}^{*}$}                      & \textbf{$\mathbf{68.2}^{**}$}                              & \textbf{$\mathbf{64.8}^{*}$}                     & \textbf{69.7}  
\\ \bottomrule                          
\end{tabular}
 \end{center}
\end{table}

\begin{table}[t]
\caption{Total training time (low-compute, low-epoch regime: 100 random models, 3 tuning epochs) on the GLUE datasets using the T5-base pretrained backbone model with 8 A100 GPUs from  $\mathcal{S}_0$ to $\mathcal{S}_1$.}\label{Tab:training-time}
\begin{center}
\small
\begin{tabular}{cccccccc}
\toprule                             \textbf{SST-2}              & \textbf{MNLI}                & \textbf{QNLI}                & \textbf{QQP}                 & \textbf{RTE}                & \textbf{STS-B}               & \textbf{MRPC}                & \textbf{CoLA }               \\ \midrule 
18 mins & 22 mins & 20 mins & 40 mins & 8 mins & 12 mins & 8 mins & 6 mins \\
 \bottomrule    
\end{tabular} 
\end{center}
\end{table}

\section{On the Discovery Sequence}
In this work, we follow the discovery sequence of ``grouping patterns -- trainable parameter allocation -- tunable groups -- strategy assignment'': 
\begin{enumerate}
    \item To explore and understand the design patterns in all the layers in large pre-trained models in scale, it is necessary and more efficient to study the layers in the unit of groups. So we start with the grouping patterns.
    \item Once figuring out the optimal grouping patterns, it is then important to explore how to allocate the trainable parameters to these different groups in order to study more subtle designs with fair comparisons (e.g., this would allow comparing different patterns of strategy assignments without the impact from different trainable parameters.).
    \item Next, it becomes influential to examine which groups need to be learned during fine-tuning before we dig into the strategy assignment patterns. Because it is only meaningful to study assigning strategies to different groups after we figure out which groups need to be learned.
    \item Finally, we study the tuning strategy assignment, which is the most subtle design.
\end{enumerate}

\end{document}